\documentclass[10pt,journal,compsoc]{IEEEtran}
\ifCLASSOPTIONcompsoc
  \usepackage[nocompress]{cite}
\else
  \usepackage{cite}
\fi
\ifCLASSINFOpdf
\else
\fi
\usepackage{times}
\usepackage{epsfig}
\usepackage{graphicx}
\usepackage{amsmath}
\usepackage{amssymb}
\usepackage{mathtools}
\usepackage{enumitem}
\usepackage{caption}
\usepackage{booktabs}
\usepackage{xcolor}
\usepackage{blindtext}
\newcommand{\mbf}{\mathbf}
\allowdisplaybreaks

\def\med{\text{med}}

\begin{document}
%
\title{Robust Isometric Non-Rigid Structure-from-Motion}

%
%
%
%

\author{Shaifali~Parashar,~Daniel Pizarro and Adrien Bartoli 
\IEEEcompsocitemizethanks{{\IEEEcompsocthanksitem S. Parashar did this work with EnCoV-Institut Pascal-CNRS/Universit\'e Clermont Auvergne, Clermont-Ferrand, France.
\protect\\
\IEEEcompsocthanksitem D. Pizarro is with
GEINTRA, Universidad de Alcal\'a, Alcal\'a de Henares, Spain.
\protect\\
\IEEEcompsocthanksitem A. Bartoli is with EnCoV-Institut Pascal- CNRS/Universit\'e  Clermont Auvergne, Clermont-Ferrand, France.}
}}

\IEEEtitleabstractindextext{%
\begin{abstract}
Non-Rigid Structure-from-Motion (NRSfM) reconstructs a deformable 3D object from keypoint correspondences established between monocular 2D images. 
Current NRSfM methods lack statistical robustness, which is the ability to cope with correspondence errors.
This prevents one to use automatically established correspondences, which are   prone to errors, thereby strongly limiting the scope of NRSfM.
We propose a three-step automatic pipeline to solve NRSfM robustly by exploiting isometry.
Step \textit{(i)} computes the optical flow from correspondences, step \textit{(ii)} reconstructs each 3D point's normal vector using multiple reference images and integrates them to form surfaces with the best reference and step \textit{(iii)} rejects the 3D points that break isometry in their local neighborhood. 
Importantly, each step is designed to discard or flag erroneous correspondences.
Our contributions include the robustification of optical flow by warp estimation, new fast analytic solutions to local normal reconstruction and their robustification, and a new scale-independent measure of 3D local isometric coherence.
Experimental results show that our robust NRSfM method consistently outperforms existing methods on both synthetic and real datasets.
\end{abstract}

}


  \maketitle

\begin{figure*}
\centering
    \includegraphics[width=.8\textwidth]{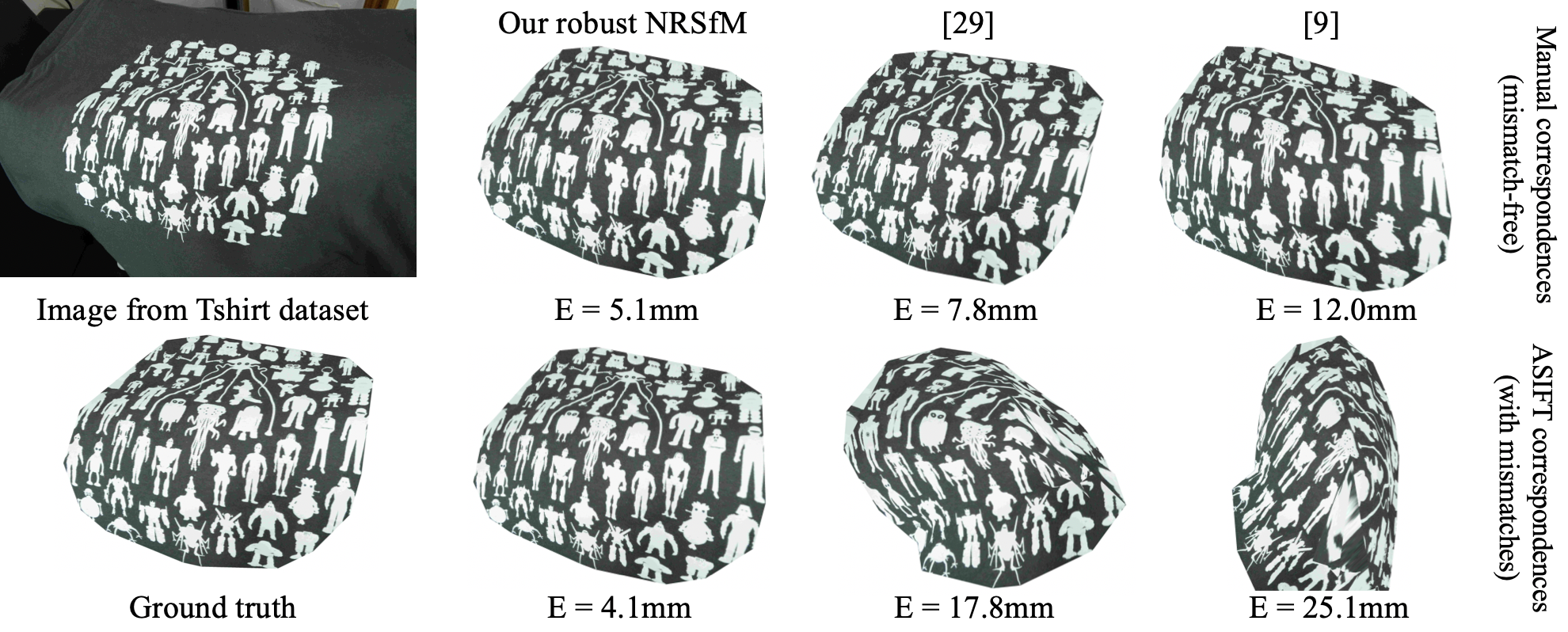}
    \caption{Reconstruction of an image from the Tshirt dataset using {\em automatic correspondences}. This dataset was introduced in~\cite{Ajad2014} with {\em manually clicked correspondences}. With these `perfect' correspondences, both~\cite{Ajad2014,Parashar2016} achieved good 3D reconstruction quality.  However, they both fail when the correspondences are obtained automatically using ASIFT, because these contain mismatches. Our robust method, on the other hand, handles both the manual and automatic correspondences well.}
    \label{fig:fig_intro}
\end{figure*}%

\IEEEdisplaynontitleabstractindextext

%
\IEEEpeerreviewmaketitle

\IEEEraisesectionheading{\section{Introduction}\label{sec:introduction}}
\label{sec:intro}
The 3D reconstruction of a deformable object from monocular images is a key computer vision problem to which NRSfM forms a promising approach.
NRSfM takes keypoint correspondences as inputs and reconstructs a 3D point cloud for each image. 
The correspondences are typically obtained from either temporally-organized, short-baseline images extracted from a video or temporally-unorganized, wide-baseline images. 
Most NRSfM methods are for short-baseline data~\cite{DelBue2004,Dai2014,Gotardo2011,Torresani2008,
Bregler2000} as the wide-baseline case was studied later~\cite{Ajad2014,Ajad2017,Varol2009}.
Most of the recent methods exploit isometry, which has been established as a widely applicable object deformation model~\cite{Ajad2017,Parashar2017,Parashar2019,Russell2014,Varol2009}.
The recent methods have brought considerable improvements in accuracy and computation time~\cite{Ajad2017,Parashar2017,Gotardo2011,Dai2014,Russell2014}.
However, the majority of existing methods are non-robust: they may tolerate noise in image point positions up to some level but do not cope with erroneous correspondences. 
This strongly limits their scope as any automatic correspondence computation method is likely to make mistakes.
This is mitigated in the short-baseline case, where correspondences may be obtained from optical flow methods such as~\cite{Sundaram2010,Garg2013} and successfully used in NRSfM. 
Failure will however occur when the amplitude of correspondence drift, which grows with the number of images, reaches the maximal noise level tolerance of NRSfM.
The wide-baseline case is however far worse, for it relies on establishing correspondences, using for instance SIFT~\cite{Lowe2004}, which contain far more errors than optical flow and a large amount of missing data.

A robust NRSfM method must cope with erroneous correspondences by classifying each image point as an inlier or an outlier while performing the reconstruction. 
It is important to classify each image point and not each correspondence, as a correspondence spanning multiple images is generally erroneous due to a few points only but still provides useful constraints.
Developing a robust NRSfM method is extremely challenging, mainly because the constraints one may use are much weaker than rigidity in SfM~\cite{Hartley2008} and SLAM~\cite{davison07a}.
The high level of robustness reached in the rigid case is largely based on the principle of random sampling, as popularized by RANSAC~\cite{fischler81a}.
As deformation is a local phenomenon, this principle can unfortunately not be used in NRSfM. 

We propose a three-step robust NRSfM pipeline for isometric deformations (see figure~\ref{fig:fig_intro}).
The idea is to leverage our correspondence-wise solution to NRSfM~\cite{Parashar2017}, which is based on the local optical flow derivatives. 
Such a solution is less vulnerable to  errors as it does not use the entire set of correspondences together at the early stages of reconstruction. 
However,~\cite{Parashar2017} falls in the category of NRSfM methods~\cite{Ajad2014,Varol2009,Vicente2012,Taylor2010} that use a single reference image to perform reconstruction, which may substantially degrade the performance.
Each step in our robust NRSfM pipeline is carefully designed to ensure stability, statistical robustness and rejection of the inconsistent image points:
{\em (i)} {\it Computation of the optical flow derivatives at the correspondences.} This step interpolates the correspondences using robust warp fitting and differentiation.
{\em (ii)} {\it Up-to-scale correspondence-wise reconstruction.} This step reconstructs the surface normal at each 3D point independently for each correspondence using multiple reference images. It keeps the normals corresponding to the best reference image, which is the one that yields the most coherent normals. It finally integrates the normal field to obtain up-to-scale 3D point clouds. 
{\em (iii)} {\it Isometry consistent filtering and rescaling.} This step filters out the 3D points incoherent with their neighborhood according to isometry, finds the relative scale between the 3D point clouds and rescales them.

Beyond the proposed pipeline, we address several important problems required to implement its steps, leading to the following four contributions.
Our first contribution, in step {\em (i)}, is a robustification of Schwarps~\cite{Pizarro2016}, a stable but non-robust warp estimation method. 
Our second and third contributions, in step {\em (ii)}, are respectively a tremendous acceleration and a robustification of the normal reconstruction method~\cite{Parashar2017}.
This method takes a correspondence with its local differential structure as input and reconstructs the surface normal for each image.
The first problem with~\cite{Parashar2017} is that it is prohibitively slow due to the use of a computationally expensive generic polynomial system solver~\cite{Henrion2003} for multivariate equations.
Relatively faster polynomial solvers such as~\cite{Kukelova2012} cannot be used as they need both linear and non-linear constraints, some of which independent of the image observations.
On the other hand, solving univariate equations has been known to be extremely fast and reliable.
Our second contribution is to convert the multivariate equations in~\cite{Parashar2017} to a single univariate equation which guarantees a very fast solution. 
We propose two ways to obtain the univariate equation: by substitution and by resultants.
The method of substitution does not require additional tools to transform the equations and is therefore extremely fast. However, it does not guarantee a real solution.
The method of resultants is time-consuming when directly applied. We propose an offline computation which leads to a gain in speed of two orders of magnitude compared to~\cite{Parashar2017} and brings it almost on par with the method of substitution.
Importantly, the method of resultants guarantees at least one real solution.
The second problem with~\cite{Parashar2017} is that it is not robust, as it gives the same weight to all images and relies on an arbitrary selected reference image.
Our third contribution is a robust estimation method based on splitting the input images into small subsets.
For each subset, we select the reference image as the image giving the most coherent normal reconstruction, as measured with a robust statistic, and discard the incoherent points.
Our fourth contribution, in step {\em (iii)}, is a scale-independent isometric consistency measure, which we use to reject the locally inconsistent 3D points.

We experimentally compared our method with existing ones on synthetic and real datasets, showing that it leads to more accurate results with and without correspondence errors, while coping with beyond $50\%$ correspondence errors, an unprecedented achievement in NRSfM.

\section{Previous Work}
The first NRSfM method was probably~\cite{Bregler2000}, which introduced the low-rank shape model that imposes  the time-varying 3D shape to follow a linear combination of shape bases. 
Numerous improvements were then proposed, including non-linear refinement~\cite{DelBue2004}, spatial smoothness~\cite{Torresani2008} and a quadratic deformation model~\cite{Fayad2009}.
Inspired by the shape bases, the trajectory basis model imposes that each point trajectory follows a pre-computed basis~\cite{Akhter2009}.
This was improved by using a DCT basis to cope with large deformations~\cite{Gotardo2011}.
\cite{Dai2014} proposed a convex relaxation of the shape bases model and solved it using convex optimization. Recently,~\cite{Agudo2016} expressed NRSfM using forces acting on the shape and a force basis.
In parallel, template-based methods were developed successfully~\cite{Bartoli2015,Salzmann2011,Brunet2011}, achieving robustness due to the stronger constraints provided by the template~\cite{Collins2015,Ngo_2015_ICCV}. 
The success of these methods inspired the development of NRSfM with physics-based deformation models, in particular isometry~\cite{Parashar2017,Ajad2014,Varol2009,Taylor2010,Vicente2012,Russell2014}. ~\cite{Ajad2017,ji17a} proposed an NRSfM method using the inextensibility relaxation of isometry. In contrast to most of the existing methods which use the calibrated perspective or orthographic camera,~\cite{Parashar2018,Probst2018} solve NRSfM with the uncalibrated perspective camera.~\cite{Probst2018} exhaustively searches for a focal length which maintains the best possible isometric consistency of reconstruction across views.~\cite{Parashar2018} computes resultants to eliminate the surface from first- and second-order constraints, leaving only the focal length as unknown.

However, the constraints available in NRSfM are much weaker than in SfM and in  template-based reconstruction. Thus, NRSfM methods have not yet reached the same level of robustness, most of them being highly sensitive to correspondence errors. Some however have made a step towards robustness. 
\cite{golyanik2017} discards correspondences by testing compatibility with a shape prior. The need for a shape prior however strongly limits its applicability. 
\cite{Varol2009} constrains the shape by learnt linear local models and \cite{Ajad2017} minimises the $L_1$ norm of slack variables modeling the deviation between the observed and predicted points.
In practice, \cite{Varol2009} breaks at about $10\%$ and \cite{Ajad2017} at about $20\%$ correspondence errors. 
The other NRSfM methods, which do not have a robust design, break at a few percents of correspondence errors, except~\cite{Dai2014}, which we have found to  break at $20$-$25\%$.
In contrast, our robust NRSfM method uses a pipeline where each step was  carefully designed to handle correspondence errors and breaks beyond $50\%$ correspondence errors.

\section{Fast Solutions to Isometric NRSfM}
\label{sec:fast}
\begin{figure}
    \centering
    \includegraphics[width=0.4\textwidth]{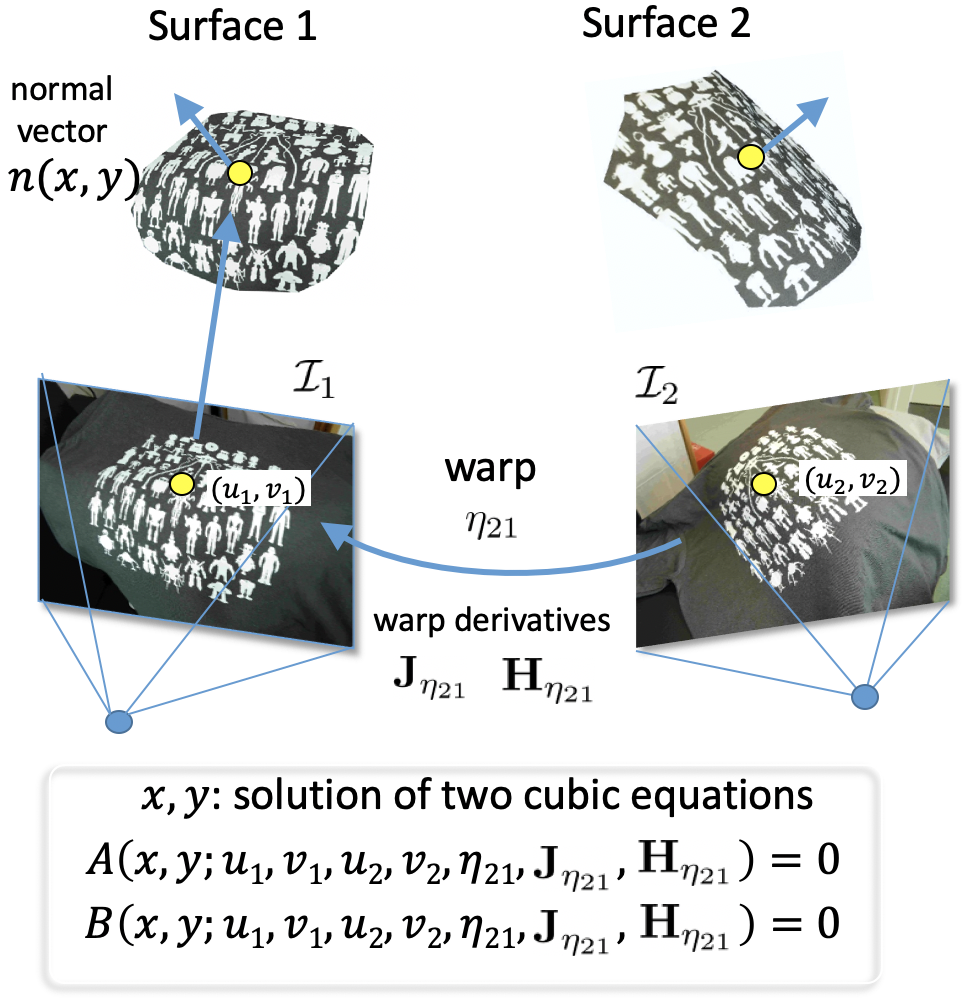}
    \caption{Notation, illustrated in a two-view NRSfM example.}
    \label{fig:diag}
\end{figure}

 We consider a set of $M$ images of a deforming object with point correspondences. We fix
a reference image of index $i = 1$ and evaluate
the $\eta_{ji}$ warps, $j\in[2,M]$, using Schwarps~\cite{Pizarro2016}. Schwarps penalise the residual of the  Schwarzian equations, which are second-order PDEs whose solutions are homographies. Schwarps thus preserve the local projective structure of the image transformation, leading to better performance than other solutions based on penalising the warp's second-order derivatives. Using $\eta_{ji}$, we hallucinate point correspondences by creating a grid on the reference image and transferring its points to the other images.
Our goal is to
retrieve the depth and normal for each of these points on each image. 
Following~\cite{Parashar2017}, the essential first step is to reconstruct the normal in the reference image, from the bi-cubic reconstruction equations. 
The normal and depth for all images are then easily found and can be jointly refined by iterative nonlinear minimisation.
The reference normal reconstruction problem being non-convex, having a reliable initial estimate is essential. 
The generic polynomial optimiser~\cite{Henrion2003} was used in~\cite{Parashar2017} to minimise the sum of squares of the bicubic reconstruction equations, which results in extremely slow performance.  
In contrast, our new solutions rely on solving univariate polynomials, which is known to be very fast and stable. Our first method uses substitution to construct the univariates from the reconstruction equations, whereas our second approach uses the theory of resultants.
We first recall the reconstruction equations from~\cite{Parashar2017} and then present our
two fast ad hoc initialisation solutions. 
Our notation is illustrated by figure~\ref{fig:diag}.
\subsection{Reconstruction Equations}
 We now explain how the isometric reconstruction equations from~\cite{Parashar2017} are derived. These equations require concepts from Riemmanian differential geometry, mainly the metric tensor and the Christoffel Symbols (CS), to model the differential relationship between isometric surfaces and their image projections.

 We describe a surface using an embedding function $\mathbf{P}=\varphi(\mathbf{p})$, where $\mathbf{p}=(u,v)^\top$ and $\mathbf{P}=(P_x, P_y, P_z)$. We assume that $\varphi$ is at least twice differentiable. The surface tangent plane is generated by the embedding's $3\times 2$ Jacobian matrix $\mathbf{J}$:
\begin{equation}
\mathbf{J} = \dfrac{\partial \varphi}{\partial \mbf{p}} = \begin{pmatrix}\dfrac{\partial \varphi}{\partial u} & \dfrac{\partial \varphi}{\partial v}  \end{pmatrix}.
\label{eq:jacobian}    
\end{equation}
Specifically, the tangent plane at any point $\mathbf{p}$, is the linear subspace generated by the column vectors of $\mathbf{J}$.
The normal vector $\mathbf{n} = \frac{\partial \varphi}{\partial u} \times \frac{\partial \varphi}{\partial v}$ of the tangent plane coincides with the surface normal.
Locally, the tangent plane induces a metric to measure differential distances between points on the surface. This metric is captured by the metric tensor $\mathbf{g}$, also known as the first fundamental form, given by:
\begin{equation}\mathbf{g} = \mathbf{J}^\top. \mathbf{J}\end{equation}
The metric tensor is a $2\times 2$ quadratic form which depends on the surface's embedding. For a change of coordinates ${\mathbf{p}}=\eta (\mathbf{\bar{p}})$, the surface embedding in the new coordinates $\bar{\varphi}=\varphi \circ \eta $ has the following metric tensor:
\begin{equation}\mathbf{\bar{g}} = \mathbf{J}^\top_\eta \mathbf{g} \mathbf{J}_\eta \quad\mbox{with}\quad \mathbf{J}_\eta = \dfrac{\partial \eta}{\partial \mathbf{\bar{p}}}.\end{equation}
The CS of the second kind, $\mathbf{\Gamma}^u$ and $\mathbf{\Gamma}^v$, are $2\times 2$ matrix functions that measure the local rate of change in the metric tensor. They are thus defined from the first-order derivatives of $\mathbf{g}$, which involve up to second-order derivatives of $\varphi$. The CS define the surface curvature and its geodesics. They also admit a change of variable that requires the first and second derivatives of $\eta$. We refer the reader to \cite{DoCarmo2016} and our original method developed in \cite{Parashar2017} for the detailed expressions of the CS in tensor notation.

One important result from~\cite{Parashar2017} is the invariance of the metric tensor and the CS under isometric deformations of the observed surface. Given two isometric surfaces (\textit{i.e.}, surfaces related by an isometric mapping), defined by their embeddings $\varphi_1$ and $\varphi_2$, the metric tensors and the CS are preserved:
\begin{equation}
\mathbf{g}_1 = \mathbf{g}_2 \qquad \mathbf{\Gamma}_1^u = \mathbf{\Gamma}_2^u \qquad \mathbf{\Gamma}_1^u =\mathbf{\Gamma}_2^u.
\label{eq:metric_preservation}
\end{equation}
The constraints in equation~\eqref{eq:metric_preservation} allow us to find the NRSfM solution for a pair of views $\mathcal{I}_1,\mathcal{I}_2$ from the warp function $\eta_{21}$. For each view $i\in\{1,2\}$ we define a surface embedding:
\begin{equation}
\varphi_i(\mathbf{p}_i) =\frac{1}{\beta_i(\mathbf{p}_i)}(\begin{matrix}\mathbf{p}_i&1\end{matrix})^\top,
\label{eq:image_embedding}
\end{equation}
where $\mathbf{p}_i=(u_i,v_i)$ are normalized image coordinates, obtained by multiplying pixel coordinates by the inverse of the intrinsic matrix. The function $\beta_i(\mathbf{p})$ is the inverse depth function. Using $\beta$ instead of the depth function greatly simplifies our equations. The Jacobian matrix of $\varphi_i$ is expressed as:
\begin{equation}
  \mathbf{J}_i = \dfrac{1}{\beta}\left(\begin{matrix}1-u_i x_i & -v_i y_i \\ -v_i x_i & 1-v_i y_i \\ -x_i & -y_i\end{matrix}\right),  
\end{equation}
where $(x_i,y_i) = \dfrac{1}{\beta_i}\left(\frac{\partial \beta_i}{\partial u_i},\frac{\partial \beta_i}{\partial v_i}\right)$.  Importantly, $(x_i,y_i)$ are the local reconstruction unknowns for view $\mathcal{I}_i$, since they allow us to recover the surface normal as:
\begin{equation}
    \mathbf{n}=\begin{pmatrix}x_1&y_1&1-x_1 u_1-y_1 v_1\end{pmatrix}^\top.
\label{eq:normal}
\end{equation}
By identifying $(u_2,v_2)=\eta_{21}(u_1,v_1)$ as a change of variable between $\mathcal{I}_1$ and $\mathcal{I}_2$, and by imposing the constraints of equation~\eqref{eq:metric_preservation}, we find the reconstruction equations.
First, the local isometric deformation constraint in terms of metric tensors is:
\begin{equation}
\mathbf{g}_2 = \mathbf{J}_{\eta_{21}}^\top\mathbf{g}_1\mathbf{J}_{\eta_{21}},
    \label{eq:NRSFMeq1}
\end{equation}
where $\mathbf{g}_i=\mathbf{J}^\top_i \mathbf{J}_i$. 
 The expressions on both sides of equation \eqref{eq:NRSFMeq1} are symmetric $2\times 2$ matrices and therefore yield 3 equations. The known and unknown quantities are $(u_1,v_1,u_2,v_2,\mathbf{J}_{\eta_{21}})$ and $(\beta_1,x_1,y_1,\beta_2,x_2,y_2)$ respectively.  $(\beta_1,\beta_2)$ are cancelled by taking ratios of the 3 equations, which leaves 2 reconstruction equations for an image pair with $(x_1,y_1,x_2,y_2)$ as unknowns.
Second, the CS between the two surfaces are preserved up to a change of variable given by $\eta_{21}$, from which \cite{Parashar2017}~derived the following linear relationship between $(x_1,y_1)$ and $(x_2,y_2)$: 
\begin{equation}
\left(\begin{matrix}x_2 \\ y_2\end{matrix} \right) = \mathbf{J}_{\eta_{21}}^\top\left(\begin{matrix}x \\ y\end{matrix}\right) -  \left(\begin{matrix}0&1\\1&0\end{matrix}\right)\mathbf{J}_{\eta_{21}}^{-1}\left(\begin{matrix}h_3 \\ h_4\end{matrix}\right),
    \label{eq:NRSFMeq2}
\end{equation}
where $\left(\begin{smallmatrix}h_3 \\ h_4\end{smallmatrix}\right)$ is the second column of $\mathbf{H}_{\eta_{21u}}$, the Hessian matrix of $\eta_{21}$ with respect to $u$. Substituting equation~\eqref{eq:NRSFMeq2} in equation~\eqref{eq:NRSFMeq1}, we arrive at 2 cubic equations with only $(x_1,y_1)$ as unknowns, that for simplicity we refer to as $(x,y)$ hereinafter. The cubics are:
\begin{align}
\nonumber A &=   a_{30} x^3 + a_{21} x^2y + a_{12} xy^2 + a_{03} y^3  + a_{20} x^2 \\    &+ a_{11} xy 
+ a_{02} y^2 + a_{10} x + a_{01} y + a_{00}, \label{eq:eq1}\\ 
\nonumber B &=   b_{30} x^3 + b_{21} x^2y + b_{12} xy^2 + b_{03}y^3 + b_{20} x^2 \\  &+ b_{11} xy  + b_{02} y^2 + b_{10} x + b_{01} y + b_{00}, \label{eq:eq2}
\end{align}
where the coefficients $a_{kl}$ and $b_{kl}$  are expressed in terms of  $\mathbf{p}_1$, $\mathbf{p}_2$ and:
\begin{equation}
\nonumber
 \mathbf{J}_{\eta_{21}}=\left(\begin{smallmatrix}
j_1 & j_3 \\ j_2 & j_4
\end{smallmatrix}\right) \qquad
\mathbf{H}_{\eta_{21u}}=\left(\begin{smallmatrix}
h_1 & h_3 \\ h_2 & h_4
\end{smallmatrix}\right) \qquad 
\mathbf{H}_{\eta_{21v}}=\left(\begin{smallmatrix}
h_3 & h_5 \\ h_4 & h_6
\end{smallmatrix}\right).
\end{equation}
Appendix~\ref{appendixA} shows the complete expressions of these coefficients.
According to B\'ezout's theorem, this system has up to $9$ solutions, so $3$ images at least are needed to ensure a single solution.
By fixing a reference image, a system of $2(M-1)$ cubics~\eqref{eq:eq1} and~\eqref{eq:eq2} can be constructed for a correspondence over $M>2$ images with only 2 variables.
  By solving for $(x,y)$, we obtain the surface normals using equation~\eqref{eq:normal}. The local depth is then obtained by integrating the local normals over the entire surface.



We use the system of cubics $A$, $B$ in $(x,y)$, as~\cite{Parashar2017}. The system can be assembled efficiently; the remaining challenge is to solve it efficiently.~\cite{Parashar2017} minimises the sum of squares of cubics for all image pairs using the generic polynomial solver~\cite{Henrion2003}. In addition to being utterly computationally expensive, such a solution strategy may lead to severely degraded results due to just a single erroneous image pair, because least-squares are not statistically robust. Erroneous image pairs occur when the geometry is ill-conditioned or when the amount of erroneous correspondences is overly important. In contrast, the proposed method solves the cubics by pairs, by converting them to univariate polynomials which are easily solvable. This has a double advantage: first, it is much faster, and second, it allows us to find a statistically robust consensus amongst all cubics. The proposed method thus deals with erroneous image pairs, covering a large extent of deficiencies which may happen in practice. Concretely, the erroneous image-pairs are identified by detecting the high residuals they cause for most solutions and discarded. Ruling out such image-pairs leads to a solution that is not affected by even highly erroneous image-pairs.
In summary, we propose strategies to solve the system for each image pair separately. This gives multiple solutions for a single normal and we pick the one that satisfies the majority of the image-pair constraints using a statistically robust criterion. This is a simple yet efficient way to identify and discard erroneous image-pairs, while obtaining a fast solution to the system.

\subsection{Fast Solution using Substitution}
\label{sec:fs}
\noindent{\bf Deriving the univariates.}
We propose a change of variable from $(x,y)$ to $(z_1\,z_2)^\top=\mathbf{J}^\top_{\eta_{21}}(x\,y)^\top$ leading to considerably simplified cubics:
\begin{align}
\nonumber A' &\!=\!  c_{21} z_1^2z_2\!+\! c_{12} z_1z_2^2  \!+\!c_{20} z_1^2 
 \!+ \!c_{11} z_1z_2\!+\! c_{02} z_2^2\!+\! c_{10} z_1 \\ &\!+ \!c_{01} z_2\!+\! c_{00},\label{eq:eq1_tr}
\\   B' &\!=\!   z_1C
\! + \!d_{03}z_2^3 \!+\! d_{02} z_2^2  \!+\! d_{01} z_2 \!+\! d_{00}, \label{eq:eq2_tr}
\\
\nonumber &
\text{where } 
C  \!=\! d_{12} z_2^2 \!+\! d_{11} z_2 \!+\! d_{10} \text{ and:} \\ \nonumber 
c_{21} & \!=\! 2e_1(e_2t_2\! -\! v_2)\! +\! 2e_2e_4, \text{ } 
c_{12}  \!=\! 2e_1(u_2\! -\! e_2t_1)\! -\! 2e_2e_3, 
\\ \nonumber 
c_{20}  &\!=\! e_1e_5\!-\!e_2(j_3^2\!+\!j_4^2), \text{ } 
c_{11}  \!=\! 4(e_2t_1\!-\!u_2)e_4\!+\!4(v_2\!-\!e_2t_2)e_3, \\ \nonumber 
c_{02}  &\!=\! e_2(j_1^2\!+\!j_2^2)\!-\!e_1e_6,  \text{ } 
c_{10} \! = \! 2(u_2\!-\!e_2t_1)(j_3^2\!+\!j_4^2)\!-\!2e_5e_3 , \\ \nonumber 
c_{01}  &\!=\! 2e_6e_4\!-\!2(v_2\!-\!e_2t_2)(j_1^2\!+\!j_2^2) , \\ \nonumber  
c_{00}  &\!=\! e_5(j_1^2\!+\!j_2^2)\!-\!e_6(j_3^2\!+\!j_4^2) , d_{00}  \!=\! e_5e_7\!+\!e_8(j_3^2\!+\!j_4^2),\\ \nonumber 
d_{12}  &\!=\! e_2e_4\!-\!e_1(v_2\!-\!e_2t_2), \text{ } 
d_{11}  \!=\! e_1e_5\!-\!e_2(j_3^2\!+\!j_4^2), \\ \nonumber 
d_{10}  &\!=\!(v_2\!-\!e_2t_2)(j_3^2\!+\!j_4^2) \!-\!e_5e_4
, \text{ } 
d_{03} \! =\! e_1(u_2\!-\!e_2t_1)\!-\!e_2e_3, \\ \nonumber 
d_{02}  &\!=\! e_2e_7\!+\!e_1e_8\!-\!2(u_2\!-\!e_2t_1)e_4\!+\!2(v_2\!-\!e_2t_2)e_3, \\ \nonumber 
d_{01}  &\!= \!\!-\!e_5e_3\!-\!2(v_2\!-\!e_2t_2)e_7\!+\!(u_2\!-\!e_2t_1)(j_3^2\!+\!j_4^2)\!-\!2e_8e_4, \\ \nonumber 
e_1 &\!=\! 1 \!+\! u_1^2 \!+\! v_1^2, \text{ } e_2 \!=\! 1 \!+\! u_2^2 \!+\! v_2^2,    \\ \nonumber 
e_3 &\!=\! j_1u_1\!+\!j_2v_1, \text{ }e_4 \!=\! j_3u_1\!+\!j_4v_1, \text{ } e_5\! = \!1\!-\!2t_2v_2\!+\!e_2t_2^2,  \\ \nonumber  e_6 &\!= \!1\!-\!2t_1u_2\!+\!e_2t_1^2, 
e_7 \!=\!j_1j_3\!+\!j_2j_4,  e_8 \!=\!t_2u_2\!+\!t_1v_2\!-\!e_2t_1t_2,\\ \nonumber 
 \text{ }t_1 &\!=\! \!-\!(j_3h_3\!+\!j_4h_4), \text{ } 
t_2 \!=\! \!-\!(j_1h_3\!+\!j_2h_4). 
\end{align}
Note that even though $x,y$ are shared across all images, $z_1,z_2$ are specific to the image pair related by the warp derivatives $j_1,j_2,j_3,j_4$.
We observe that  equation~\eqref{eq:eq2_tr} is linear in $z_1$. 
We first assume $C \neq 0$, substitute $z_1$ from equation~\eqref{eq:eq2_tr} in equation~\eqref{eq:eq1_tr} and rewrite $A'$ as:
\begin{align}
\label{eq:eq_final}
\nonumber 
A'' &\!=\! (c_{21}z_2\!+\!c_{20})(d_{03}z_2^3\!+\!d_{02}z_2^2\!+\!d_{01}z_2\!+\!d_{00})^2\!  \\ \nonumber &+\!(c_{02}z_2^2\!+\!c_{01}z_2 \!+\! c_{00} ) C^2 
\\ & \nonumber
\!-\! z_2^2(c_{12}z_2^2 \!+\! c_{11}z_2 \!+\! c_{10})(d_{03}z_2\!+\!d_{02}\!) C 
\\
& \!-\! (c_{12}z_2^2 \!+\! c_{11}z_2 \!+\! c_{10})(\!d_{01}z_2\!+\!d_{00}) C. 
\end{align}
On expanding, $A''=0$ yields a sextic (a degree 6 polynomial) in $z_2$. Computing the roots of a sextic is known to be simple, stable and computationally cheap
~\cite{Basu2016}.
We now consider $C = 0$, in which case $B'$ in equation~\eqref{eq:eq2_tr} becomes a univariate cubic in $z_2$, which can be easily solved. Substituting $z_2$ in equation~\eqref{eq:eq1_tr} then yields a quadratic equation in $z_1$, which can also be easily solved. 

\noindent \textbf{Solving.} We have two possible cases depending on the nullity of $C$. Numerically, an  explicit choice is not desirable. Hence we solve both cases and collect the solutions in a potential solution set.
Selecting the optimal solution is done by using extra images, as described in the next paragraph. We now analyse the solvability of $z_1,z_2$ in each case.
The original shape variables $x,y$ are then found as $(x,y)^\top = \mbf{J}_{\eta_{21}}^{-\top}(z_1,z_2)^\top$.

In the case $C= 0$, $z_2$ is found from a univariate cubic, and hence always has at least one real solution. Using $z_2$ thus obtained, $z_1$ is found by solving a quadratic. Theoretically this does not guarantee a real solution for $z_1$. 


In the case $C\neq0$, the sextic $A''$ factors into a quadratic and a quartic as:
\begin{align}
\nonumber
 &A'' \!=\!  D^2(s_1z_2 \!-\! s_5 ) 
 \!+\!  D(s_{13}z_2^2\!-\!s_{12}z_2 \!-\! s_{11})^2  \!+\! \\ 
&D(s_{13}z_2^2\!-\!s_{12}z_2 \!-\! s_{11})(s_{10}z_2^2\!+\!s_{9}z_2 \!+\! s_{8})
 \label{eq:eq_final_fac}
\\
\nonumber &
\text{where } 
D \!=\!s_{16}z_2^2\!+\!s_{15}z_2 \!+\! s_{14} \text{ and:}\\ \nonumber
s_1 &\!=\! 2e_2e_{11}\!-\!2e_1e_9, \text{ } 
s_5  \!=\! e_2e_{14}\! -\! e_1e_5, s_8 \!=\! 2e_3e_5\! -\!2e_{14}e_{10}, 
\\ \nonumber 
s_9 &\!=\! 4e_{10}e_{11}\!-\!4e_3e_9, \text{ } 
s_{10}  \!=\! 2e_2e_3\! -\! 2e_1e_{10}, s_{11} \!=\! e_4e_5\! -\!2e_{14}e_{9},
\\ \nonumber 
s_{12}  &\!=\! e_{14}e_2\!-\!e_1e_5\!+\!2e_9e_{12},  \text{ } 
s_{13} \!=\! e_2e_{11}\!-\!e_1e_9\!+\!e_2e_{12} ,   \\ \nonumber
s_{14}  &\!=\! e_{13}e_5\!-e_{14}e_6,   \text{ } 
s_{15}  \!=\! 2e_{11}e_{6}\!-\!e_9e_{13}, \text{ }  
s_{16}  \!=\! e_2e_{13}\!-\!e_1e_6,\\ \nonumber 
s_{2}  &\!=\! s_{10}s_{11}\!+\!s_9s_{12}\!-\!s_{13}s_8, \text{ } 
s_{3}  \!=\! s_8s_{12}\!+\!s_9s_{11}, \\ \nonumber 
s_{4}  &\!=\!s_9s_{13}-s_{10}s_{12}, \text{ } 
s_{6} \! =\! s_{12}^2-2s_{11}s_{13}, \text{ }
s_{7}  \!=\! s_{15}^2 + 2s_{14}s_{16}, \\ \nonumber 
e_{9}  &\!= v_2\!\!-\!e_2t_2,  \text{ }
e_{10}  \!= u_2\!\!-\!e_2t_1,  \text{ }
e_{11}  \!= j_2u_1\!\!+\!j_4v_1, \\ \nonumber 
e_{12} &\!=\! j_2u_1\!-\!j_3u_1, \text{ }
e_{13} \!=\! j_1^2\!+\!j_2^2, \text{ } e_{14}\! =\! j_3^2\!+\!j_4^2. 
\end{align}
The discriminant of $D$ is always non-negative.  Hence, at least one real solution is always achievable for $z_2$.  This holds even in the limit case of an infinitesimal local motion, as backed up by our experiments. 
Further, $z_1$ can be obtained by back-substituting in the linear equation~\eqref{eq:eq2_tr}.  The system has 6 possible solutions. In most cases, we find only one or two real solutions.

\subsection{Fast Solution using Resultants}
\label{sec:fr}
\noindent{\bf Deriving the univariates.}
We use the theory of resultants~\cite{Akritas1993,cox2006}, which states that the resultant of two equations that bear at least one common root evaluates to zero. 
Given two univariate polynomials of degrees $n$ and $m$ respectively, $p = \sum_{f=0}^n p_f z^f$ and $q = \sum_{f=0}^m q_f z^f$, the resultant is defined as the determinant of the Sylvester matrix, an $(n+m)\times (n+m)$ matrix formed using the coefficients of $p$ and $q$.
For instance, with $(n,m)=(2,1)$ the Sylvester matrix is:
\begin{equation}
\label{eq:syl}
S = \begin{pmatrix}
 p_2 & p_1 & p_0 \\
 q_1 & q_0 & 0  \\
0 &  q_1 & q_0  
\end{pmatrix}.
\end{equation}
In our case, the Sylvester matrix $S$ of the cubics $A$ and $B$ constructed to eliminate variable $y$ is:
\begin{align}
\label{eq:sylvester}
\small
S &= \begin{pmatrix}
a_{03} & \sigma_5 & \sigma_3 & \sigma_1 & 0 & 0 \\
0 & a_{03} & \sigma_5 & \sigma_3 & \sigma_1 & 0 \\
0 & 0 & a_{03} & \sigma_5 & \sigma_3 & \sigma_1 \\
b_{03} & \sigma_6 & \sigma_4 & \sigma_2 & 0 & 0 \\
0 & b_{03} & \sigma_6 & \sigma_4 & \sigma_2 & 0 \\
0 & 0 & b_{03} & \sigma_6 & \sigma_4 & \sigma_2 \\
\end{pmatrix}, \text{ where:} \\
  \nonumber
\sigma_1 &= a_{30}x^3 + a_{20} x^2 + a_{10}x + a_{00}, \hspace{1pt} 
\sigma_4 = b_{21} x^2 + b_{11}x + b_{01},\\  \nonumber
\sigma_2 &= b_{30}x^3 + b_{20} x^2 + b_{10}x + b_{00}, 
\hspace{5pt}
\sigma_5 = a_{02} + a_{12}x,\\  \nonumber
\sigma_3 &= a_{21} x^2 + a_{11}x + a_{01},  \hspace{38pt}
\sigma_6 = b_{02} + b_{12}x.
\end{align} 
We form $\det(S) = 0$ symbolically, which gives a univariate polynomial $P$ in $x$ of degree $9$. 

\noindent \textbf{Solving.} 
At runtime, the coefficients of $P$ are computed from the coefficients of the bivariate cubics $A$ and $B$, $x$ is obtained by solving $P=0$ and $y$ is  obtained by back-substituting $x$ in $A=0$ or $B=0$.
The degree 9 guarantees at least one real solution to $x$. On backsubstituting $x$ in either $A$ or $B$ yields cubics in $y$ which also guarantees at least one real solution to $y$.
In practice, only $1$ or $3$ solutions for $x$ and $y$ are real.
The solvability of the cubics $A$ and $B$ depends on the structure of the Sylvester matrix. Some coefficients of  $A$ and $B$ may be less significant than others in some cases, which may lead to a rank-deficient Sylvester matrix~\eqref{eq:sylvester} and thus to a wrong solution.  
It is therefore fundamental that we handle all possible cases of rank-deficiency of $S$ to ensure that a reliable solution is computed. 
 Table~\ref{tab:1} shows the Sylvester matrix for all possible combinations of degrees of $A$ and $B$. It also reports the degree of $P$ and whether it can be symbolically factored into lower degree polynomials.
In all cases, $P$ either has an odd degree or factors into odd degree polynomials, which ensures the existence of  one real root. 
Instead of computing the resultants online for each correspondence, which would takes around 3-4 seconds, we use the pre-computed analytical resultants for each of the possible 9 cases. Thus, for each correspondence, we look up for the relevant resultant based on the coefficients of $A$ and $B$. It then only takes about $1 ms$ to form the resultant.  
In practice, we choose the case at hand using the following steps:
 \begin{enumerate}
 \item Compute the medians $\lambda_a$ and $\lambda_b$ of the absolute values of the coefficients $a_{kl}$ and $b_{kl}$, $k,l\in [0,3]$ respectively.
 
 \item Discard the coefficients $a_{kl}$ and $b_{kl}$ whose absolute value falls below $\mathrm{th}_a = 0.01 \lambda_a$ and $\mathrm{th}_b = 0.01 \lambda_b$ respectively.
 
 \end{enumerate}

\begin{table}
\centering
\footnotesize
\setlength\tabcolsep{1.3pt}
\begin{tabular}{ccccc}
\toprule
Equation~\eqref{eq:eq1} & Equation~\eqref{eq:eq2} & Sylvester matrix & Degree & Factors \\
\midrule \\
Quadratic      & Cubic           &  $\begin{pmatrix}
\sigma_5 & \sigma_3 & \sigma_1 & 0 & 0 \\
0 & \sigma_5 & \sigma_3 & \sigma_1 & 0 \\
0 & 0 & \sigma_5 & \sigma_3 & \sigma_1 \\
b_{03} & \sigma_6 & \sigma_4 & \sigma_2 & 0 \\
0 & b_{03} & \sigma_6 & \sigma_4 & \sigma_2 
\end{pmatrix}$                & 9      & -       \\ \\
Linear         & Cubic           &  $\begin{pmatrix}
\sigma_3 & \sigma_1 & 0 & 0 \\
0 & \sigma_3 & \sigma_1 & 0 \\
0 & 0 & \sigma_3 & \sigma_1 \\
b_{03} & \sigma_6 & \sigma_4 & \sigma_2 
\end{pmatrix}$                & 9      & -       \\ \\
Cubic          & Quadratic       &  $\begin{pmatrix}
a_{03} & \sigma_5 & \sigma_3 & \sigma_1 & 0 \\
0 & a_{03} & \sigma_5 & \sigma_3 & \sigma_1 \\
\sigma_6 & \sigma_4 & \sigma_2 & 0 & 0 \\
0 & \sigma_6 & \sigma_4 & \sigma_2 & 0 \\
0 & 0 & \sigma_6 & \sigma_4 & \sigma_2 
\end{pmatrix}$                & 9      & -       \\ \\
Cubic          & Linear          &   $\begin{pmatrix}
a_{03} & \sigma_5 & \sigma_3 & \sigma_1 \\
\sigma_4 & \sigma_2 & 0 & 0 \\
0 & \sigma_4 & \sigma_2 & 0 \\
0 & 0 & \sigma_4 & \sigma_2 \\
 \end{pmatrix}$               & 9      & -       \\ \\
Quadratic      & Quadratic       &   $\begin{pmatrix}
\sigma_5 & \sigma_3 & \sigma_1 & 0 \\
0 & \sigma_5 & \sigma_3 & \sigma_1 \\
\sigma_6 & \sigma_4 & \sigma_2 & 0 \\
0 & \sigma_6 & \sigma_4 & \sigma_2 \\
 \end{pmatrix}$               & 8      & 5,3     \\ \\
Linear         & Quadratic       &      $\begin{pmatrix}
\sigma_3 & \sigma_1 & 0 \\
0 & \sigma_3 & \sigma_1 \\
\sigma_6 & \sigma_4 & \sigma_2 
\end{pmatrix}$            & 7      & -       \\ \\
Quadratic      & Linear          &   $\begin{pmatrix}
\sigma_5 & \sigma_3 & \sigma_1 \\
\sigma_4 & \sigma_2 & 0 \\
0 & \sigma_4 & \sigma_2 
\end{pmatrix}$               & 7      & -       \\ \\
Linear         & Linear          &     $\begin{pmatrix}
\sigma_3 & \sigma_1 \\
\sigma_4 & \sigma_2 
\end{pmatrix}$             & 5      & - \\
\bottomrule     
\end{tabular}
\\ 
\normalsize
\caption{The Sylvester matrix for the degenerate cases of cubics $A$ and $B$. The degree of the determinant of this matrix along with the degree of its possible factors are reported.}
\label{tab:1}
\end{table}
\subsection{Algorithm}  

Given a set of point correspondences over $M$ images and the $\eta_{ji}$ warps between a fixed reference frame $i$ and the other images $j\in[1,M], j\neq i$, we reconstruct the normals and depth.
The normals are found independently for each correspondence:
\begin{enumerate}
    \item \textit{Initialize $x,y$.}  
\begin{enumerate}
    \item  Randomly select a few images ($10 \%$ of $M$ or $10$, whichever is maximum), including the reference image $i$.
    \item Solve $x,y$ for each image pair $i,j$ using the substitution or resultant method. With the substitution method, obtain $z_2$ by solving equations~\eqref{eq:eq_final_fac} and~\eqref{eq:eq2_tr} and $z_1$ by substituting the real solutions for $z_2$ in equation~\eqref{eq:eq2_tr}. Set $x,y$ to $\mbf{J}_{\eta_{21}}^{-\top}(z_1,z_2)^\top$. With the resultant method, obtain a univariate equation in $x$ by expressing the resultant of $A$ and $B$ in terms of $y$ according to table~\ref{tab:1}. Using $x$, obtain $y$ by solving $A=0$ or $B=0$. 
    
    \item From the $x,y$ thus obtained, compute the sum of absolute values of $A$ and $B$ over the $M$ images. Pick the solution that gives the least residual. Flag the images with a residual larger than 10 times the median over the entire image set. 
\end{enumerate}
    
    \item \textit{Refine $x,y$.} Refine $x,y$ by minimising the sum of squares of $A$ and $B$ over the set of non-flagged images using Levenberg-Marquardt.
 \item \textit{Transfer of normals.}  $x,y$ thus obtained represent local depth derivatives in image $i$. According to~\cite{Parashar2017}, the ones on image $j$ is obtained from $(x, y)$, $\mathbf{J}_{\eta_{ji}}$ and $\mathbf{H}_{\eta_{ji}}$. The normal at each surface is finally obtained from the local depth derivatives and image coordinates. For example, for a point $u,v$ in image $i$ having $x,y$ as local depth derivatives, the normal is given by $(x,y,1-ux-vy)^\top$.

\end{enumerate}
 Finally, as in~\cite{Parashar2017},  we obtain depth by integrating the normals for each image, giving $M$ surfaces.

\begin{figure}
    \centering
    \includegraphics[width=0.48\textwidth]{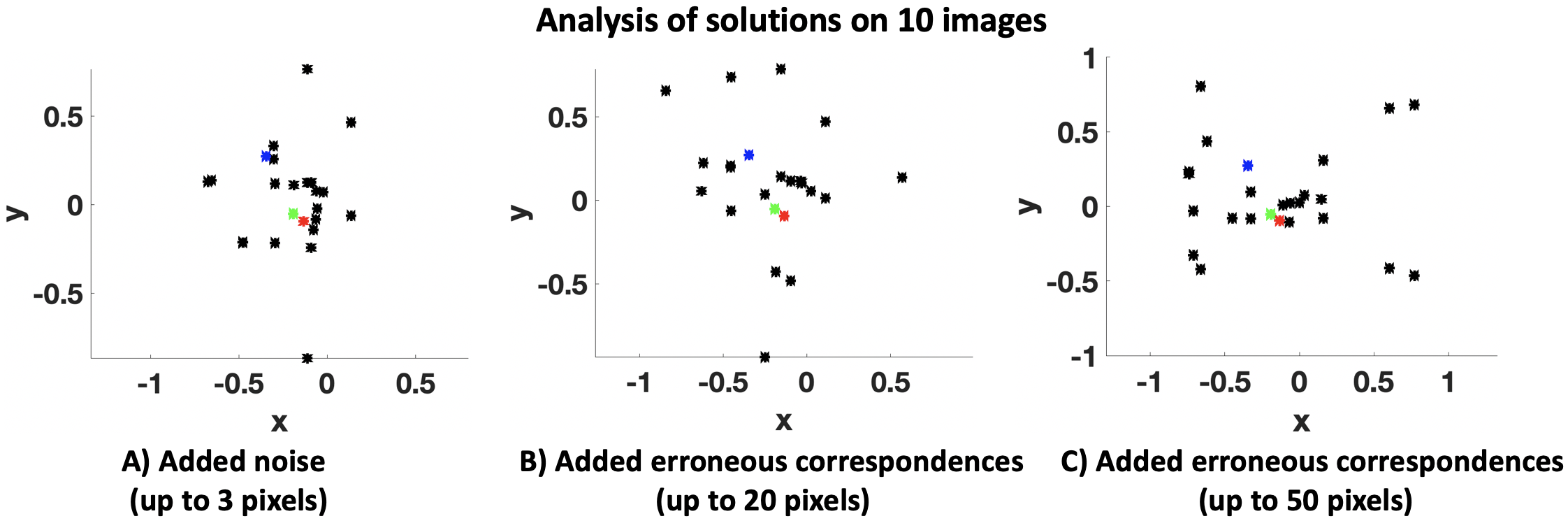}
    \caption{Solution from~\cite{Parashar2017} and our methods for three scenarios.}
    \label{fig:cases}
\end{figure}

\subsection{Comparison of~\cite{Parashar2017} and our method}
We illustrate the potential problems caused by the solution in~\cite{Parashar2017}, and how these are fixed by our method, in figure~\ref{fig:cases}. The solutions to $(x,y)$ from 10 image-pairs are shown as black dots.  Ideally, all the solutions should be identical but due to noise and correspondence errors, they are different for each image-pair. The red dot represents the true solution, the blue dot represents the solution obtained by~\cite{Parashar2017} and the green dot represents our solution. Figure~\ref{fig:cases}A demonstrates that under good registration conditions (when the image-pair registrations are all reliable), most of the solutions for all image-pairs are close to the ground truth with only slight perturbations due to noise. In this example, we added a gaussian noise of standard deviation of 3 pixels to ground truth point correspondences to simulate good -but noisy- registration conditions. The solutions are computed for 10 image-pairs on a single correspondence but we see more than 10 solutions in the plot as the pairwise solution to the cubics yields more than one solution (typically 2 or 3 solutions).
While~\cite{Parashar2017} finds a unique solution by jointly minimizing the sum-of-squares of cubics over all image-pairs, which is not robust, our strategy solves for each image-pair separately and picks a statistically robust consensus. We can observe that, even in this easy case, our solution lies closer to ground truth than~\cite{Parashar2017}'s. In contrast to figure~\ref{fig:cases}A, figures~\ref{fig:cases}B and~\ref{fig:cases}C were constructed for cases where the image-pair registrations have high errors (namely, optical drift caused by matching low textured image regions or using sparse registration techniques such as SIFT, which are prone to create erroneous correspondences) or the object moves very fast (which causes motion blur). Concretely, we created erroneous correspondences by adding random errors of up to 20 and 50 pixels respectively on $20\%$ of the point correspondences. With such errors, one can see how the solution space diversifies, which strongly affects the computation of the solution from~\cite{Parashar2017}, but not the proposed one.

\section{Proposed Robust NRSfM Pipeline}
\label{sec:pipeline}
The three main steps of our pipeline are given in the introduction.
The inputs are correspondences for $N$ images which do not need to be visible in all images.
We split the images into subsets of  $M\leq N$ images. We choose $M=7$ for wide-baseline data. For short-baseline videos, we choose $M=35$.  This is because the relative motion between two images is small in videos and a larger number of images are required to obtain well-formed constraints. However in this case, we uniformly sample the subset to pick only 7 images to be treated as reference (considering all images with small relative motion would be computationally irrelevant).
However, this number can be arbitrarily adapted. For the rest of the paper, $M=7$ means 7 images each serving as a reference for wide-baseline data and 35 images with 7 uniformly sampled references for short-baseline data.

\subsection{Step {\em (i)}: Robust Optical Flow Derivatives}
\label{sec:3.1}
The ability to reconstruct a normal field correspondence-wise at step {\em (ii)} requires that the correspondences are augmented with the optical flow derivatives at first and second orders.
Our approach is to interpolate the correspondences by computing a Schwarp, as it was shown to provide an extremely reliable estimate of these derivatives~\cite{Pizarro2016}.
We estimate a total of $M(M-1)$ Schwarps, for all pairs of images within each image subset.
The existing Schwarp computation method~\cite{Pizarro2016} is not robust.
However, image points have a local influence on the Schwarp's behavior, thanks to its high flexibility. 
We thus propose to estimate the Schwarp with~\cite{Pizarro2016} and flag as outliers those image points whose predicted position is significantly different from the measured position, indicating local inconsistency.
We then iterate these two steps until convergence, similarly to~\cite{pilet08a}.
We assume that the noise on the true image points follows a Gaussian distribution with an unknown standard deviation $\sigma$.
We use the Median Absolute Deviation (MAD) to compute an estimate $\hat\sigma= 1.4826\,\text{MAD}$ of $\sigma$ at each iteration. 
An image point is then flagged as inlier and stored in a temporary index set $\mathcal{L}$ if its position discrepancy is lower than a threshold $3\hat\sigma$, allowing us to ensure that $99.7\%$ of the true image points are kept on average.
We write the $M$-image subset as $\mathcal{I}=[1,M]$ and the set of $s$ points on image $t \in \mathcal{I}$ as $\{\mathbf{p}_t^1,\hdots, \mathbf{p}_t^s\}$. Our algorithm for an ordered image pair $(t,u) \in \mathcal{I}$ is initialized with $\mathcal{L}=[1,s]$:
\begin{enumerate}
\item 

 Compute Schwarp $\eta_{tu}$ from correspondences in $\mathcal{L}$.
 
\item 
 Predict discrepancy $d^j_u = {\left\lVert\eta_{tu}(\mathbf{p}^j_t)-\mathbf{p}^j_u\right\rVert}_{1}$ for $j\in[1,s]$ and $\hat\sigma= 1.4826\,\med_{j\in[1,s]}(d^j_u)$. 

\item 
 Flag inliers as $\mathcal{L}=\{j\in[1,s] \,|\, d^j_u < 3\hat\sigma\}$ and estimate $\hat\sigma'= 1.4826\,\med_{j\in\mathcal{L}}(d^j_u)$.

\item 
 Compute Schwarp $\eta_{tu}$ from correspondences in $\mathcal{L}$.

\item 
 If $|\hat\sigma-\hat\sigma'|<\delta$ return else loop to 2).
\end{enumerate}
The inliers are flagged during estimation to ensure robustness but all points are kept for the next step.
Convergence is assessed from the evolution of MAD, which we threshold using $\delta$, chosen as $0.1\%$ of the diagonal of image $u$. 
Note that all parameters in this step are fixed and found automatically. 

\subsection{Step {\em (ii)}: Robust Correspondence-wise Normal Reconstruction using Multiple References}
\label{sec:3.2a}

We assume that we have a {\em base} correspondence-wise normal reconstruction method, which is one of the methods given in~\S\ref{sec:fast}. 
Given a correspondence for three images or more from step {\em (i)}, this base method estimates the surface normal at all unknown 3D points of this correspondence.
The base method is not robust but serves as a key building block of our robust method described directly below.
We consider one correspondence at a time.
The robustness principle we use is to gradually remove the images which show maximal inconsistency with the others, until we settle on a consistent image subset for the given correspondence.
Concretely, we start with the complete  $\mathcal{I} = [ 1,M]$.
We thus have that initially  $\mathtt{size}(\mathcal{I})=M$ and eventually $\mathtt{size}(\mathcal{I})\leq M$, indicating the inlier image points.
Importantly, the base method requires one to fix a reference image.
We write $V_k^t$ the normal for image $k\in\mathcal{I}$ reconstructed using $t\in\mathcal{I}$ as reference image.  
For a consistent image set $\mathcal{I}$, $V^t_k$ should be relatively independent of $t$.
Therefore, the consistency of $V^t_k$ for the different reference images suggests that it was reliably estimated and thus the absence of correspondence errors.
We measure the inconsistency $S_{t,u}$ between two reference images with indices $t,u\in\mathcal{I}$ using the angle between the normal estimates they produce for all images as: 
\begin{equation}
S_{t,u} = \mathrm{angle}(V^t_k, V^u_k).
\label{eq:S}
\end{equation}
The inconsistency $U_t$ of an individual reference $t$ is then given by marginalizing $u$ in $S_{t,u}$ using the median, giving $U_t=\mathrm{median}_{u\in\mathcal{I},u\not=t} S_{t,u}$.
Finally, the overall inconsistency $G$ of $\mathcal{I}$ is given by the minimum of $U_t$ as $G=\min_{t\in\mathcal{I}}U_t$.
A small value indicates that all images were consistent and they all gave a similar reconstruction.
The consistency of an image set is decided using a threshold $\varepsilon$ on $G$. 
Our algorithm starts with $\mathcal{I} = [ 1,M]$ and iterates as:
\begin{enumerate}
\item 

 Reconstruct $V^t_k$ for $k,t\in\mathcal{I}$ using a base method in~\S\ref{sec:fast}. Find $U_t = \mathrm{median}_{u\in\mathcal{I},u\not=t} S_{t,u}$ 
  and $G=\min_{t\in\mathcal{I}}U_t$.

\item 
 If $G<\varepsilon$, set $\hat V_k$ for $k\in\mathcal{I}$ to the most consistent estimate $\hat V_k=V^t_k$ with $t=\arg\min_{t\in\mathcal{I}} U_t$ and return. If $\mathtt{size}(\mathcal{I})<5$, set $\mathcal{I}=\emptyset$ and return. Drop $t=\arg\max_{t\in\mathcal{I}} U_t$ from $\mathcal{I}$, loop to 1).

\end{enumerate}
The outputs are $\mathcal{I} \subset [1,M]$, indicating the inlier image points, and $\hat V_k$ with $k\in\mathcal{I}$, giving the reconstructed normals. In this step, only $\varepsilon$ needs to be set. We choose $\varepsilon=5$ degrees.


\subsection{Step {\em (iii)}: Robust Isometry Consistent Filtering} 
\label{sec:3.3} 
Isometric deformations preserve geodesic distances. 
Since geodesic distances are expensive to compute (they cannot be computed in closed-form for a general surface) and sensitive to noise, we approximate them with Euclidean distances, which is a fair approximation on short distances and for surfaces with mild local curvature, as shown in previous work~\cite{Ajad2017,Salzmann2011,Varol2009,Vicente2012}.
It is highly likely that erroneously reconstructed points obtained from step {\em (ii)} do not comply with isometry and can be disregarded as outliers. 
We thus measure the consistency of inter-point distances across the reconstructed point clouds to detect outliers.
However, step {\em (ii)} produces up-to-scale point clouds. In order to compare distances, we thus need to first recover the relative scales.
We use a Nearest-Neighbor Graph (NNG) as in~\cite{Ajad2017}, using the rationale that if two points are close in 3D they must also be close in the images. 
We store the indices of the $r$ nearest neighbors of each correspondence of index $k\in[1,s]$ in a vector $R_k$ of size $r$, where $s$ is the number of correspondences. We set $r = 20$. This step is also devoid of  parameters to be set manually.

\subsubsection{Robust relative scale estimation}
Without loss of generality, we set the relative scale for image 1 as $\alpha_1=1$.
We then estimate $\alpha_i$, $i\in[2,M]$.
We first define distance vectors $P_j^i$ of length $r$ computed from the reconstructed shape $i\in[2,M]$ for the neighborhood of each correspondence $j\in[1,s]$: $P_j^i(l)$ contains the 3D Euclidean distance between the points of indices $j$ and $l \in R_j$.
Each vector $P_j^i$ contributes $r$ equations towards scale estimation.
A robust estimate for $\alpha_i$ is then the median of distance ratios, given by:
\begin{equation}
\alpha_i = \med_{j \in [1,s],l \in [1,r]} \frac{P_j^i(l)}{P_j^1(l)},
\label{eq:alpha}
\end{equation}
where we only run over the indices of the correspondences visible in both images $1$ and $i$. 
If image $1$ does not contain sufficient correspondences with all other images, we use multiple reference images and propagate the estimated scales.
We then rescale the reconstructed shapes using $\alpha_i$.
\subsubsection{Robust isometry-based inlier/outlier classification}
We rescale the distance vectors $P_j^i$ as $Q_j^i = \alpha_i P_j^i$.
In the absence of correspondence errors, these should be identical over $i\in[1,M]$, due to the temporal consistency of isometry.
We thus compute an isometry consistency measure for each point as the proportion of neighbors and images for which the distance variation is lower than a threshold, which we chose as $10\%$ of the average distances in the neighborhood.
This tolerance is important to model noise and the fact that the geodesic distance was approximated by the Euclidean distance, which underestimates it.
If the computed proportion is more than $50\%$, then the point is compatible with over half of its neighbors from over half of the images where they are visible.
This suggests than the point should be classified as an inlier and as an outlier otherwise.
\begin{figure}
\centering
\includegraphics[width=0.24\textwidth]{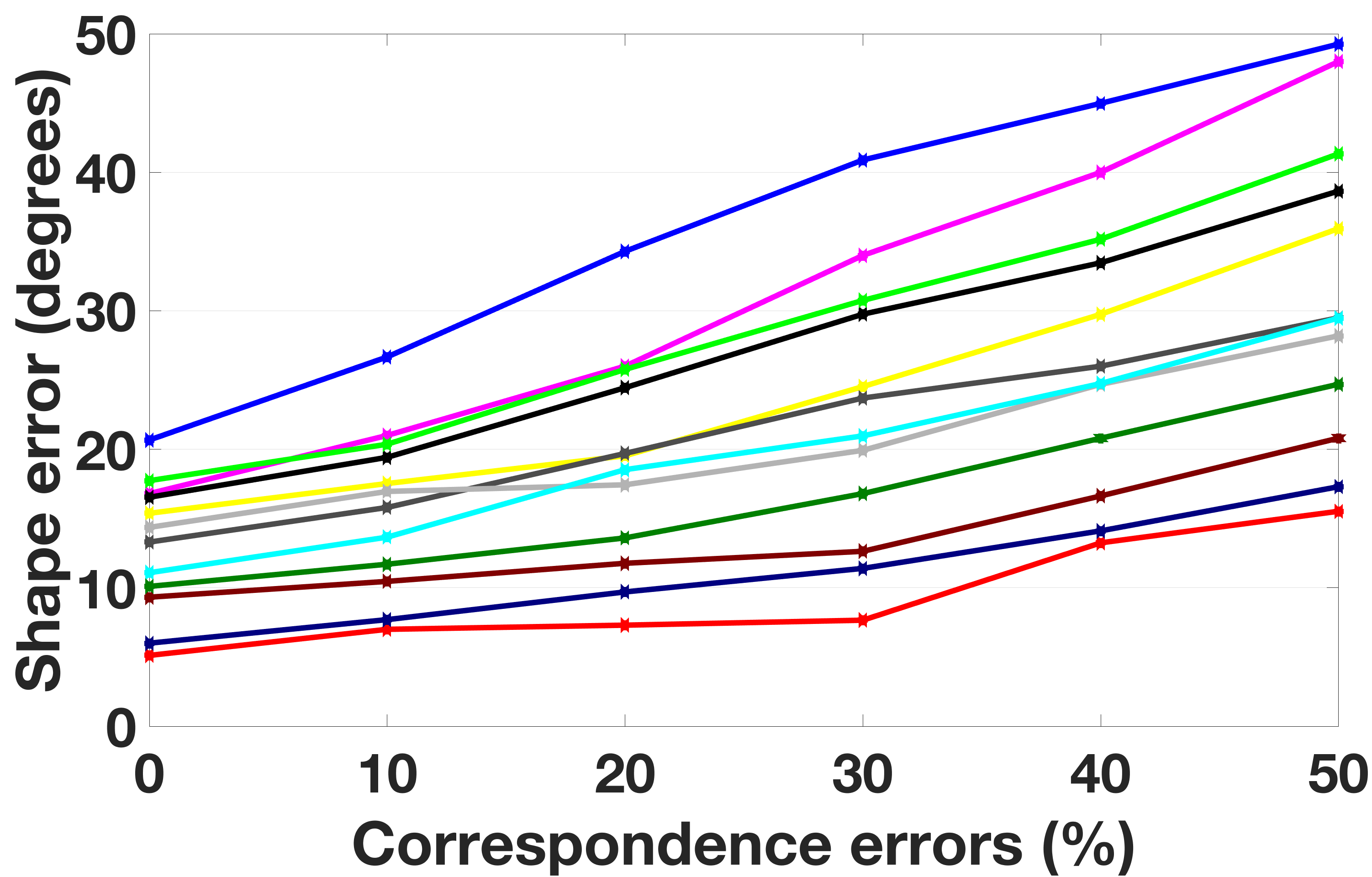}
\includegraphics[width=0.24\textwidth]{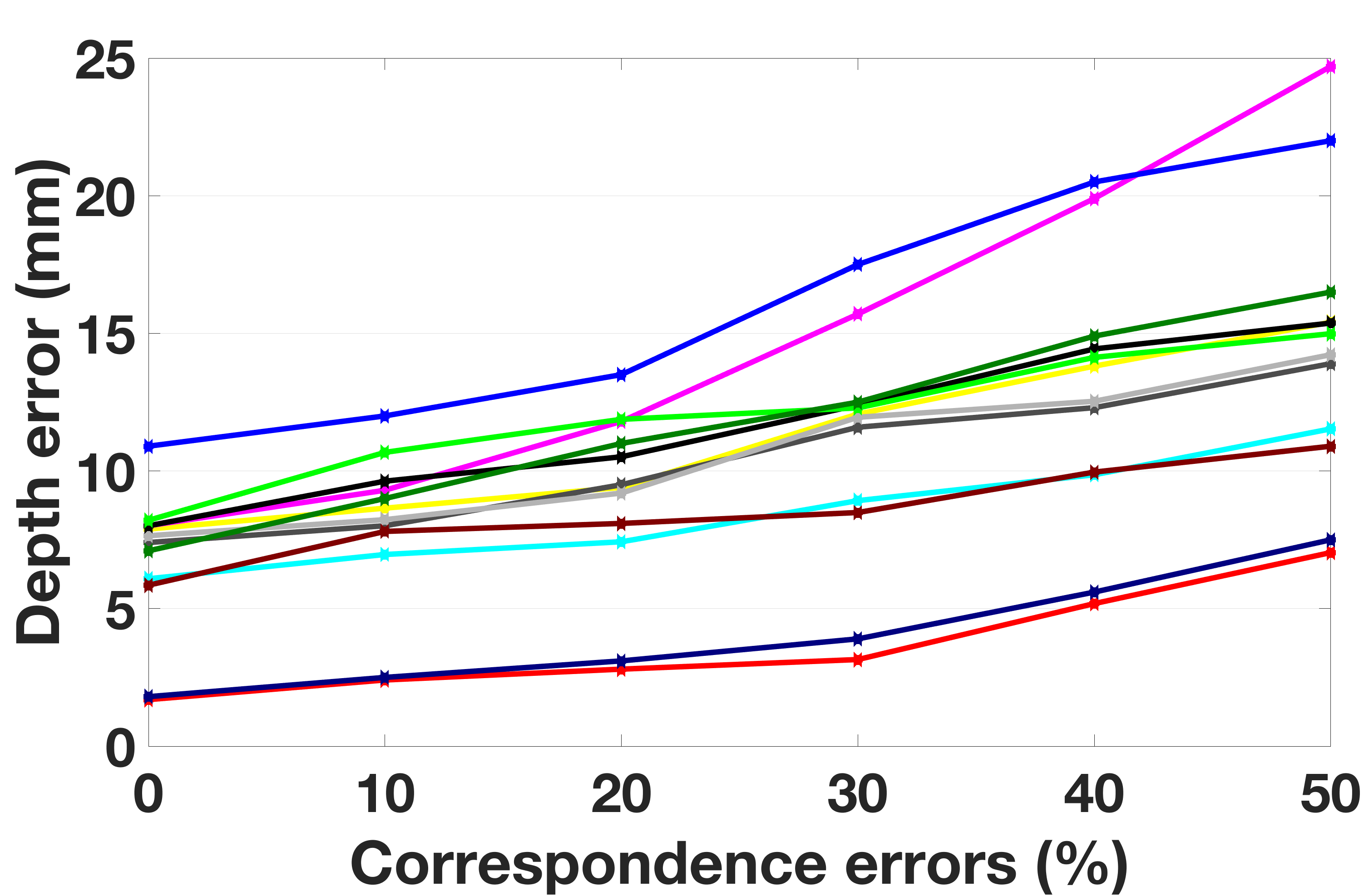}
\includegraphics[width=0.4\textwidth]{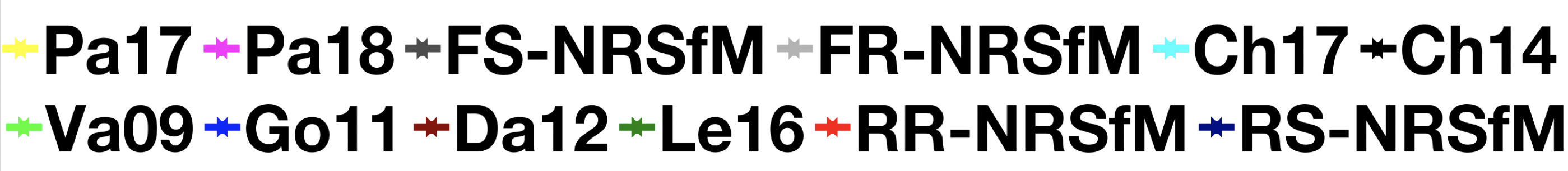}
\includegraphics[width=0.5\textwidth]{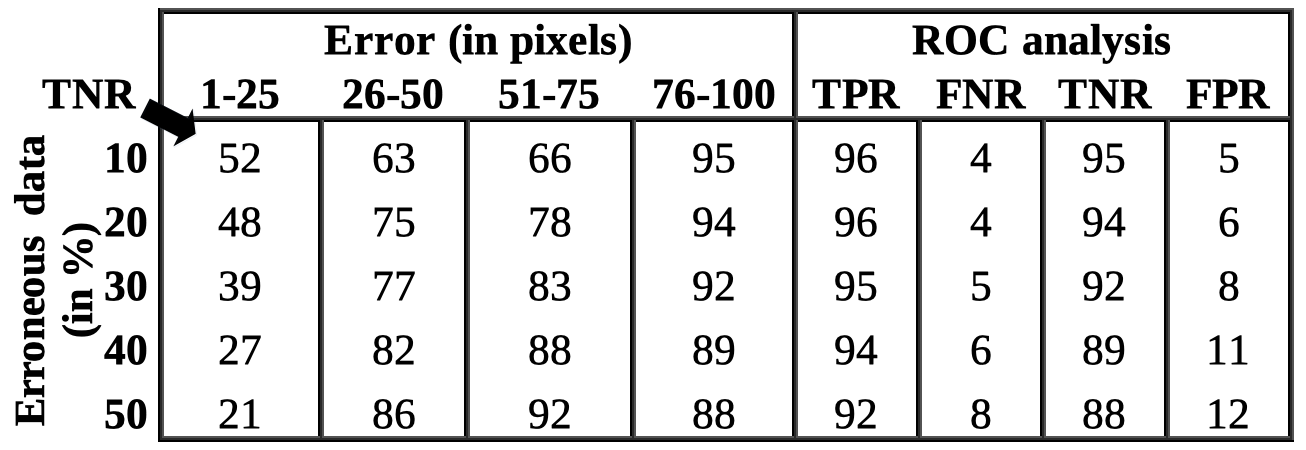}
\includegraphics[width=0.4\textwidth]{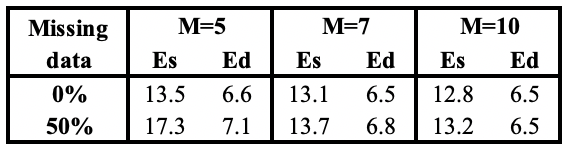}
\caption{The \textit{Cylinder} dataset. (top) Shape and depth errors ($400$ points, $7$ images, $1$ pixel noise, correspondence errors between $0-50\%$ at $100$ pixels). (middle) Analysis of \textbf{RR-NRSfM}: TNR shown across perturbation amplitudes and ROC analysis in the error range  of $76-100$ pixels. (bottom) Shape (Es) and depth (Ed) error for \textbf{FR-NRSfM} with various sizes of $M$ with/without the influence of missing data. The mean  computation time for $M$ as 5, 7 and 10 are 2.5s, 3s and 5s respectively. The performance becomes stable for $M\geq7$.}
\label{fig:2}
\end{figure}
\section{Experimental Results}
\label{sec:experiments}
\noindent \textbf{Datasets and methods.}
We used one synthetic and four real existing datasets showing various objects deforming isometrically.  
We introduce a video of a use case example depicting a realistic scenario of two objects, a paper and a cushion, recorded using Kinect.
We also introduce a footage downloaded from Youtube showing a spotted eagle-ray (an endangered species of fish).
We denote the fast local normal estimator of \S\ref{sec:fs} as \textbf{FS-NRSfM} and the one of \S\ref{sec:fr} as \textbf{FR-NRSfM}. It is the baseline reconstruction method for our robust pipeline that has three steps, as described in \S\ref{sec:pipeline}.
Step \textit{(i)} in \S\ref{sec:3.1} identifies potential mismatches and  is denoted as \textbf{MAD}. 
Step \textit{(ii)} in \S\ref{sec:3.2a} performing reconstruction from multiple reference views using \textbf{FS-NRSfM} is denoted as \textbf{MVS} and using \textbf{FR-NRSfM} as \textbf{MVR}.
Step \textit{(iii)} in \S\ref{sec:3.3} identifies 3D points as inliers and outliers based on the isometric constraint and  is denoted as \textbf{OR}.
Our proposed robust NRSfM method is denoted by \textbf{RS-NRSfM}, which is \textbf{MAD+MVS+OR} or \textbf{RR-NRSfM}, which is \textbf{MAD+MVR+OR}. We also show results of all possible combinations of these steps in order to quantify their importance.
We compare our methods with the state-of-the-art methods designed for short-baseline data, \textbf{Go11}~\cite{Gotardo2011}, \textbf{Da12}~\cite{Dai2014} and  \textbf{Le16}~\cite{Lee2016}, wide-baseline data, \textbf{Ch17}~\cite{Ajad2017}, \textbf{Ch14}~\cite{Ajad2014}, \textbf{Va09}~\cite{Varol2009} and \textbf{Pa17}~\cite{Parashar2017}, which deals with both short and wide-baseline data. We also compare with \textbf{Pa18}~\cite{Parashar2018}, which solves NRSfM with uncalibrated cameras on both short and wide-baseline data. In order to keep the comparison fair we however use the known intrinsics in  \textbf{Pa18}. 
 
\textbf{RS-NRSfM} and \textbf{RR-NRSfM} are the only methods that perform inlier-outlier classification. We identify the true correspondences as positives P and the false ones as negatives N.
TP and TN represent the image points which are correctly classified  as true and false correspondences respectively. Similarly, FP and FN represent the image points which are incorrectly classified  as true and false correspondences respectively. We  compute the true positive rate TPR = TP/P, true negative rate TNR = TN/N, false positive rate FPR = 1-TNR and false negative rate FNR = 1-TPR. 

\noindent \textbf{Measured Errors.} 
We measure the depth  error (RMSE  between reconstructed and ground truth 3D points in mm) and the  shape error (RMSE between reconstructed
and  ground truth  normals in  degrees).
The depth error is position dependent and reflects the extrinsic quality of the reconstruction. 
It depends on the object size.
A depth error lower than $5\%$ of the object size indicates a reasonable reconstruction.
However, a flawed reconstruction may also yield a low depth error by lying in the vicinity of the ground truth.
The shape error thus complements the depth error by being position independent and reflecting the intrinsic quality of the reconstruction. 
A reasonable reconstruction is obtained for a shape error lower than $20$ degrees. 
Overall, a reconstruction is successful if depth and shape errors are below $5\%$ of the object size and $20$ degrees respectively.
\begin{figure*}
\centering
\includegraphics[width=\textwidth]{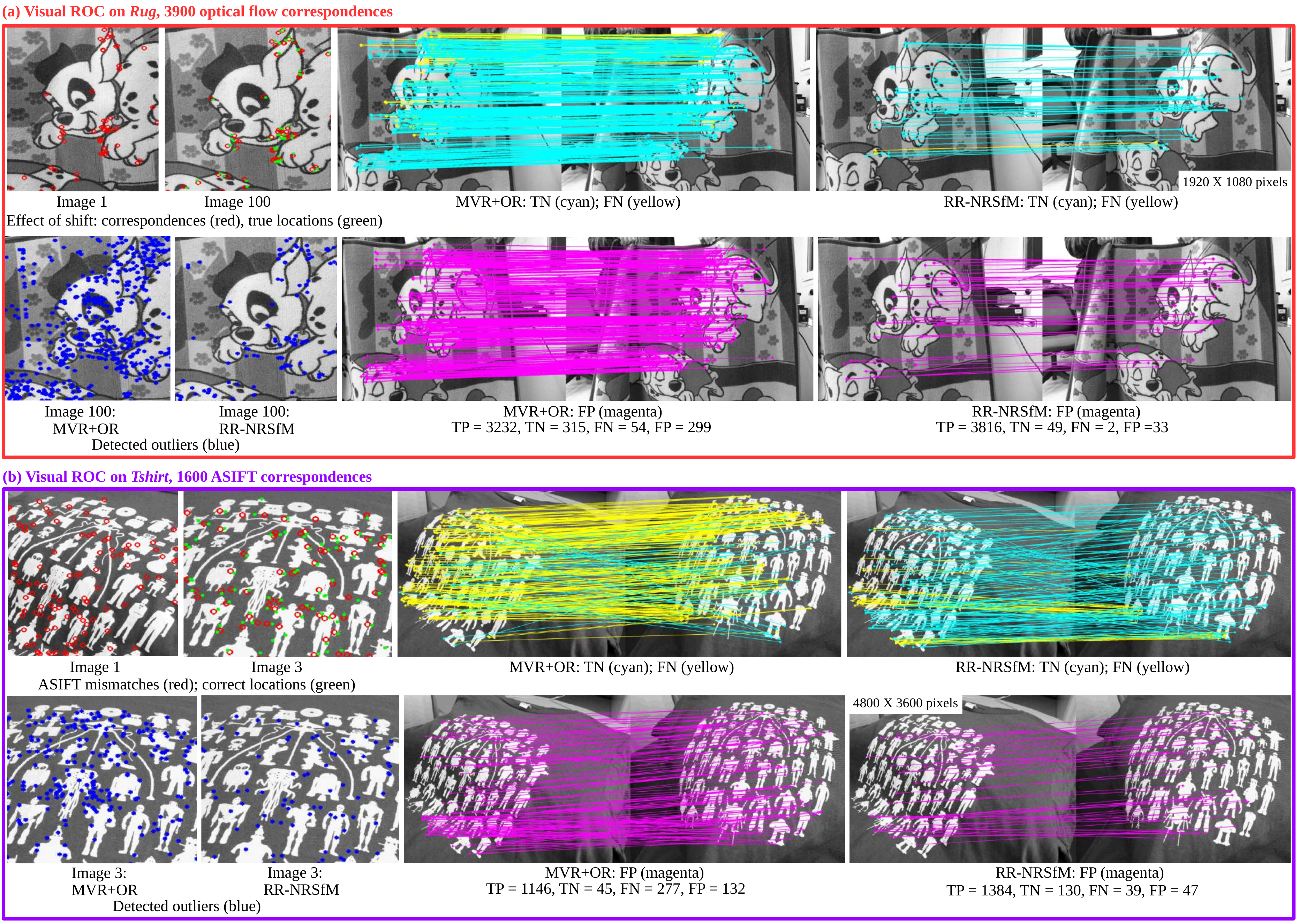}
\caption{Visual ROC for \textit{Rug} and \textit{Tshirt}. \textbf{RR-NRSfM} performs \textbf{MAD+MVR+OR}. The use of \textbf{MAD} and \textbf{OR} drastically reduces the falsely classified 3D points. In \textit{Rug},  \textbf{OR} identifies less points when used with \textbf{MAD} while in \textit{Tshirt}, it identifies more incorrect 3D points.}
\label{fig:4}
\end{figure*}

\subsection{Synthetic Dataset}
We used the \textit{Cylinder} dataset~\cite{Parashar2017}.
We generated $1920\times1080$ images of a cylindrical surface of length $20$ cm deforming isometrically. 
We compared the methods on  $N=7$ images  with $400$ correspondences.
We introduced a Gaussian noise of $1$ pixel standard deviation and between $0-50\%$ correspondence errors by applying a uniformly distributed perturbation with $100$ pixels standard deviation.
The results are shown in figure~\ref{fig:2} (top). 
All methods are sensitive to correspondence errors and degrade when the number of correspondence errors increases. 
Our robust methods \textbf{RS-NRSfM} and \textbf{RR-NRSfM} perform consistently well, even with correspondence errors: their shape error is consistently lower than $15$ degrees and their depth error lower than $10$ mm, which represents $5\%$ of the object size. 
Note that although~\cite{Parashar2017} is based on the same system of equations than ours, it finds a non-robust solution and uses only one of the images as reference. In contrast, our robust method solves for each image-pair separately and uses several reference image candidates, picking a solution as the best consensus to a set of coherent image-pairs. This contributes to a better solution overall as, even in the absence of mismatches, the optical flow computation contains errors due to lack of features in some areas of the object or challenging imaging conditions, such as motion blur. We discuss this in more details in the upcoming section. 
\textbf{RR-NRSfM} performs slightly better than \textbf{RS-NRSfM} as its underlying local normal estimator \textbf{FR-NRSfM} performs slightly better than \textbf{FS-NRSfM}.
Our local normal estimators \textbf{FR-NRSfM} and \textbf{FS-NRSfM} show a similar performance, slightly better than \textbf{Pa17}.
However, they all break at about $20\%$ correspondence errors. \textbf{Pa18} shows a slightly worse performance than \textbf{Pa17} in the range of low correspondences errors but it breaks much earlier than \textbf{Pa17}, at about only $10\%$.
This shows that our robust pipeline brings a substantial improvement compared with the local estimators \textbf{FS-NRSfM}, \textbf{FR-NRSfM}, \textbf{Pa18} and \textbf{Pa17}.
\textbf{Ch17} performs slightly better than \textbf{FR-NRSfM} and \textbf{FS-NRSfM} but significantly worse than \textbf{RS-NRSfM} and \textbf{RR-NRSfM}.
\textbf{Da12} and \textbf{Le16} show a good performance and tolerance to correspondence errors up to about $30\%$.
\textbf{Go11} has a poor performance, even in the absence of correspondence errors.
\textbf{Ch14} and \textbf{Va09} have a marginally good performance and degrade quickly, not tolerating any correspondence errors.
Overall, we observe that our robust pipelines \textbf{RR-NRSfM} and \textbf{RS-NRSfM} bring strong benefits. 

\noindent \textbf{Local normal estimators: Pa17, Pa18, FS-NRSfM and FR-NRSfM.} \textbf{Pa17} estimates local normal by minimizing the sum of squares of bivariate cubics using an computationally expensive solver. \textbf{FS-NRSfM} simplifies this problem by reducing the bivariate cubics into sextic univariate using substitution. It is also able to identify and remove the images that do not yield good constraints. \textbf{Pa18} and \textbf{FR-NRSfM} both use resultants to segregate variables. \textbf{Pa18} uses resultants to
isolate focal length from two other variables, which represent the local shape, and finds an optimal solution from the univariate constraints. It then uses this solution to obtain univariate polynomials for the local shape variables which are solved by minimizing the respective sum-of-squares. Even when used with a known focal length, \textbf{Pa18} solves degree 18 polynomials, making it sensitive to noise and correspondence errors. In contrast, \textbf{FR-NRSfM} uses resultants to solve for local shape. It pre-computes them analytically and finds a solution to univariate polynomials of degree 9. Like \textbf{FS-NRSfM}, it is also capable of identifying and removing images that yield outlying constraints. 

\noindent \textbf{Analysis of \textbf{RS-NRSfM} and \textbf{RR-NRSfM}.}
We used the same synthetic dataset to analyze the ability of our robust methods to detect correspondence errors. They perform very much alike in terms of detecting outliers.
We measured  TNR for a varying correspondence error rate and magnitude of the point perturbation causing the errors.
We report the TNR for \textbf{RR-NRSfM} in figure~\ref{fig:2} (middle). The statistics for \textbf{RS-NRSfM} are similar and we therefore do not show them.
As expected, we observe that correspondence errors caused by a large perturbation ($76$-$100$ pixels) are very well detected.
TNR naturally degrades with decreasing perturbation magnitude.
This confirms the intuition that a large error is easier to detect than a small error, which can more easily be identified as noise by a deformation model.
ROC analysis shows very successful results, as  TPR and TNR are mostly  $90\%$ or higher.
The correspondences errors larger than 25 pixels 
(spanning up to $50\%$ data) can be detected by our proposed method with a very high accuracy, above $80\%$. However, in case of $60\%$ errors, the performance sharply degrades to $53\%$ and reduces to $20\%$ for the data with $80\%$ errors. 

\noindent \textbf{Parameter tuning.} 
Importantly, all parameters were fixed to a single value for all experiments.
We use Schwarps to obtain first and second order derivatives, which requires the schwarzian parameter to be tuned. We have fixed it to 1e-3, as suggested by~\cite{Parashar2017} and it does not need to be changed. 
Both \textbf{FS-NRSfM} and \textbf{FR-NRSfM} solve univariate equations to initialize the system. In order to ensure a good initial solution, we should solve these univariates multiple times and pick the solution which is most suitable. 
We discard image pairs whose residual (on cubics $A,B$) is 10 times higher than the median over the entire image-set. This threshold is approximately twice the optimal threshold for the gaussian noise distribution and was empirically chosen.
We found that a good initialization can be achieved by solving with $10-15$ images. 
An important parameter is $M$, the number of images in a subset on which our robust pipeline operates. Our methods \textbf{FS-NRSfM} and \textbf{FR-NRSfM} can reconstruct from only 3 images. Therefore, $M$ should be 3 at minimum but the solution with just 3 images can be highly affected by noise. In addition, the possibility of missing data should also be taken into account. Therefore, we fix $M=7$. We also made experiments with $M=5$ and $M=10$. The computation time is 2.5s, 3s and 5s for the progressive values of $M$. 
With only 5 images, the reconstruction quality is similar to 7 images, with a lower computation time. However in case of missing data which can be as high as $50\%$ in realistic scenarios, using only 5 images does not ensure constraints on enough points and the reconstruction accuracy drops significantly, see figure~\ref{fig:2} (bottom). With 7 images, the performance stabilizes with a small overhead in computation time. With 10 images, the accuracy is slightly improved but the computational overhead is not worth the increase in performance.

Our robust NRSfM pipeline has 3 steps described in \S\ref{sec:3.1}, \S\ref{sec:3.2a}
and \S\ref{sec:3.3} and each of them has parameters that need to be tuned.
Step \textit{(i)} computes optical flow derivatives and has two important parameters: $\sigma$ and $\delta$. $\sigma$ decides how many points are to be discarded at each step and it is automatically chosen using a statistical criterion. $\delta$ decides when the convergence is achieved so that we do not need to remove points any further. It is image size dependent and we thus fix it to $0.1\%$ of the image diagonal.
Step \textit{(ii)} requires $\varepsilon$. Ideally the multiple normals reconstructed using the different reference images should be close but it is not the case in realistic scenarios. Therefore, we allow the closeness threshold $\varepsilon=5$ degrees. This means that we consider an image-set as genuine if the median of the angles between reconstructed normals with different reference images is 5 degrees. In our experiments, we found that this is a generous limit to allow the reconstruction from image correspondences contaminated with Gaussian noise of up to 10 pixels. 
Step \textit{(iii)} requires $r$, which decides the neighborhood size for computing the isometric consistency. Since we deal with semi-dense data, we need $r$ to be small enough for the Euclidean approximation of geodesic distances to be valid. Also, we need to consider enough points to have a reliable estimate. We therefore set $r=20$ neighbors.

\noindent \textbf{Computation time.} 
We used a desktop with an i5 CPU and 8GB RAM.
Our base normal estimators \textbf{FS-NRSfM} and \textbf{FR-NRSfM} take about 7-10~ms to solve the reconstruction equations while \textbf{Pa17} takes about $1.5$~s for the same data.
Recall that the reconstruction equations for one correspondence depend on two variables for any number of images and are very fast to assemble (forming there coefficients takes much less than $10$~ms).
\textbf{FS-NRSfM} and \textbf{FR-NRSfM} are thus more than $100$ orders of magnitude faster than \textbf{Pa17}.
\textbf{Pa18} takes about $200$~s to compute the resultant, $80$~s to obtain the focal length and $20$~ms to find the local shape.
The MATLAB implementation of our robust pipeline \textbf{RS-NRSfM} and \textbf{RR-NRSfM} runs in about $100-120$~s for $10$ images without any code optimization and can presumably be made way faster in a parallelized implementation, as each subset of images and each correspondence could be treated independently at some level. The computation time for other methods is comparable to \textbf{RS-NRSfM} and \textbf{RR-NRSfM} for a small number of images. However, most of them have a cubic complexity  while \textbf{RS-NRSfM} and \textbf{RR-NRSfM} has a quadratic complexity, giving them a huge advantage for large sets of images.
\begin{figure*}
\centering
\includegraphics[width=0.8\textwidth]{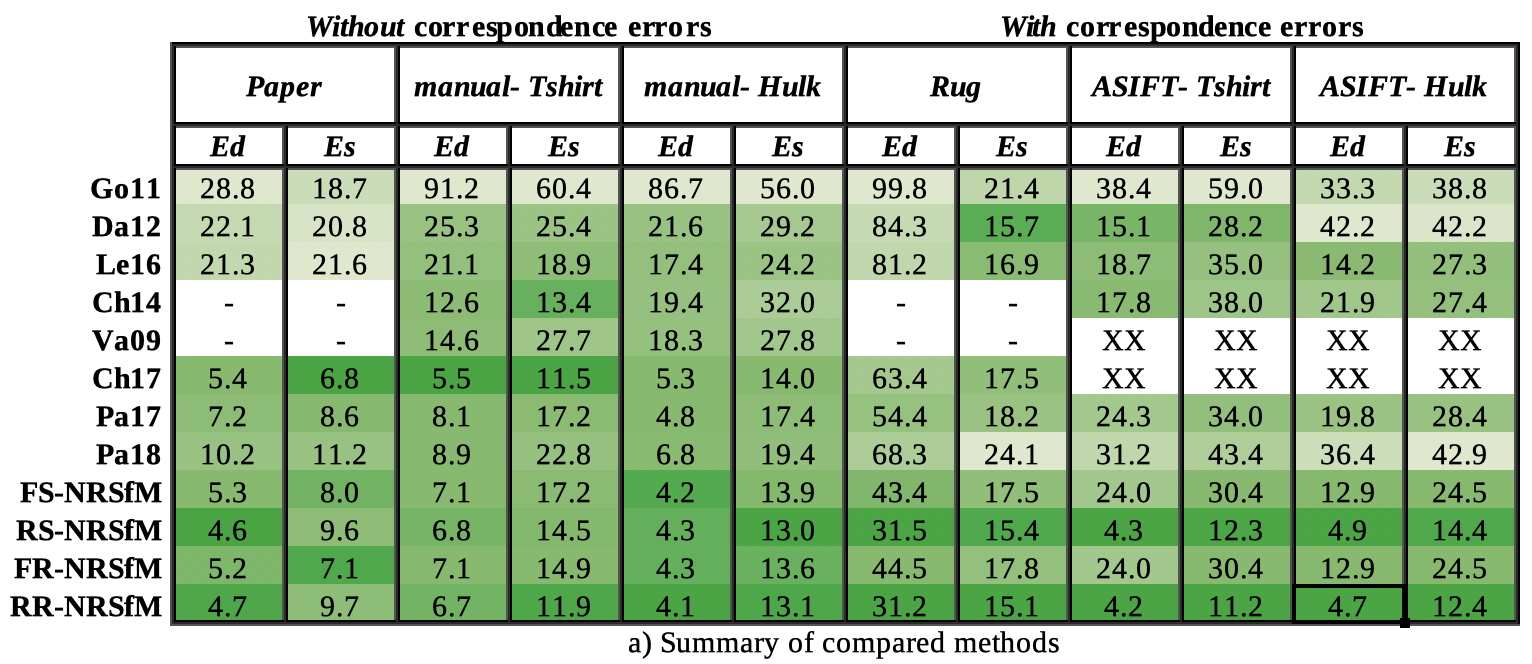}
\includegraphics[width=0.8\textwidth]{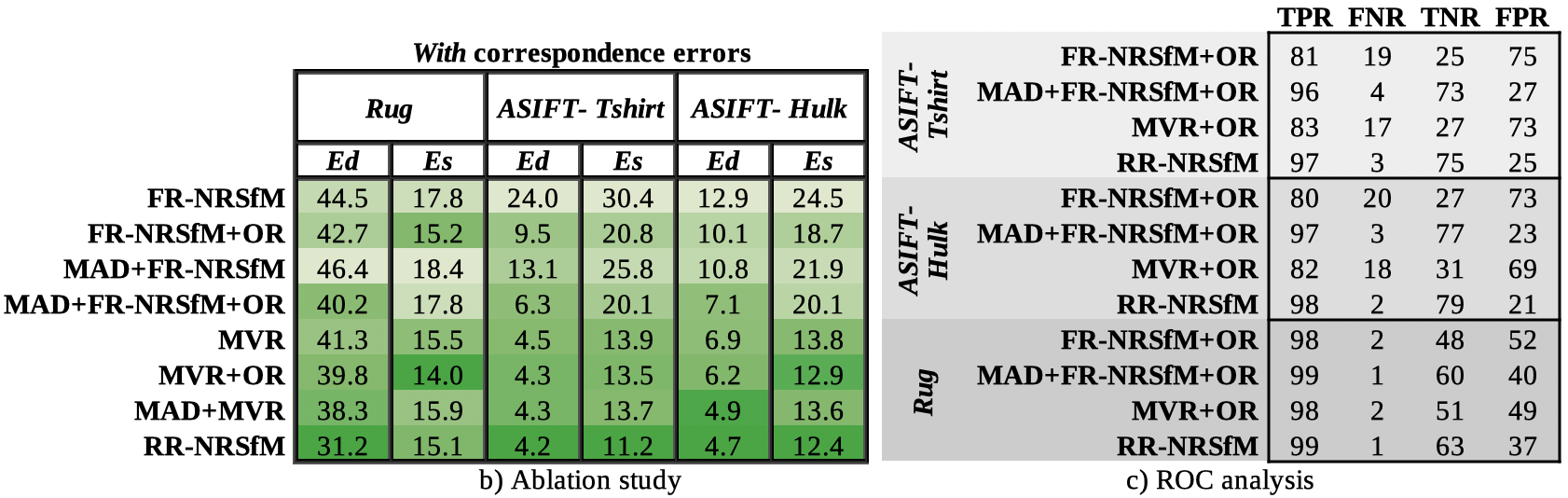}
\caption{(a) Summary of methods. For each dataset, the depth (Ed) and shape (Es) errors are reported. `--' represents the methods that did not complete in $24$ hours. `XX' represents the methods which failed due to  missing data. (b) Ablation study and (c) ROC analysis of our robust pipeline using \textbf{FR-NRSfM} as a base method.}
\label{fig:5}
\end{figure*}

\subsection{Real Datasets}
\noindent \textbf{{\em Without} Correspondence Errors: \textit{Paper}~\cite{Varol2012}, \textit{manual-Tshirt} and \textit{manual-Hulk}~\cite{Ajad2014}.} 
The \textit{Paper} dataset has $190$ short-baseline images of a $35$ cm wide deforming paper  with $1500$ error-free SIFT correspondences. The $20$ cm long \textit{Tshirt} and $30$ cm long \textit{Hulk} datasets consist of 10 and 21 wide-baseline images with $85$ and $122$ error-free manually-set correspondences.
It may seem surprising that we test our robust method on data without correspondence errors.
This however represents an important sanity check.
Non-robust methods use a strong prior, which is that all the correspondences are correct.
They thus just solve for the reconstruction.
In contrast, robust methods do not use this prior and also classify the correspondences as inliers and outliers.
Therefore, for data without correspondence errors, robust methods may underperform non-robust methods.
A very important question is to which extent this happens to our robust methods and how close the detected outlier rate is to $0\%$.

Figure~\ref{fig:5} (a) shows the results. 
Since these datasets are error-free, all methods perform well  except \textbf{Go11}, \textbf{Da12} and \textbf{Le16}. They are designed for short-baseline data only, therefore they fail on  \textit{manual-Tshirt} and \textit{manual-Hulk}. \textbf{Pa17} shows a good performance.
We could not manage to run \textbf{Va09} and \textbf{Ch14} on \textit{Paper}  because of the large number of images. \textbf{Ch14} shows a decent performance on  \textit{manual-Tshirt} but fails on  \textit{manual-Hulk}. \textbf{Va09} does not perform well on these datasets.
\textbf{Ch17} shows the best performance. It works extremely well for \textit{manual-Tshirt} and \textit{manual-Hulk}  as it requires wide-baseline data with very high perspective images  which are the characteristics of these datasets.
\textbf{Pa17} performs well, similarly to \textbf{Ch17}. \textbf{FS-NRSfM} and \textbf{FR-NRSfM} improve the results and perform even closer to \textbf{Ch17}, \textbf{FR-NRSfM} being slightly better than \textbf{FS-NRSfM}. \textbf{Pa18} performs  worse than \textbf{Pa17}.

The three steps of our robust pipeline,
 \textbf{MAD}, \textbf{MVS}/\textbf{MVR} and \textbf{OR} eliminate points from input.  This is why we see that the performance of \textbf{FS-NRSfM} and \textbf{FR-NRSfM} is either slightly degraded or only insignificantly improved when combined with \textbf{MAD} and \textbf{MV}. However, \textbf{OR} always improves the results as it removes 3D points from the reconstruction.
It removes almost $1\%$ 3D points when used without \textbf{MAD}.  On the other hand, when used with \textbf{MAD}, it slightly improves the results while removing only $0.1\%$ of the 3D points on the entire data. Hence, \textbf{MAD} prevents \textbf{OR} from removing a large number of 3D points. 
Overall, \textbf{RS-NRSfM} and \textbf{RR-NRSfM} both pass the sanity check of error-free correspondences: they obtain the best performance or are very close to the best performing non-robust method, while ruling out only $0.1\%$ of the data as outliers.

\begin{figure}
\centering
\includegraphics[width=0.47\textwidth]{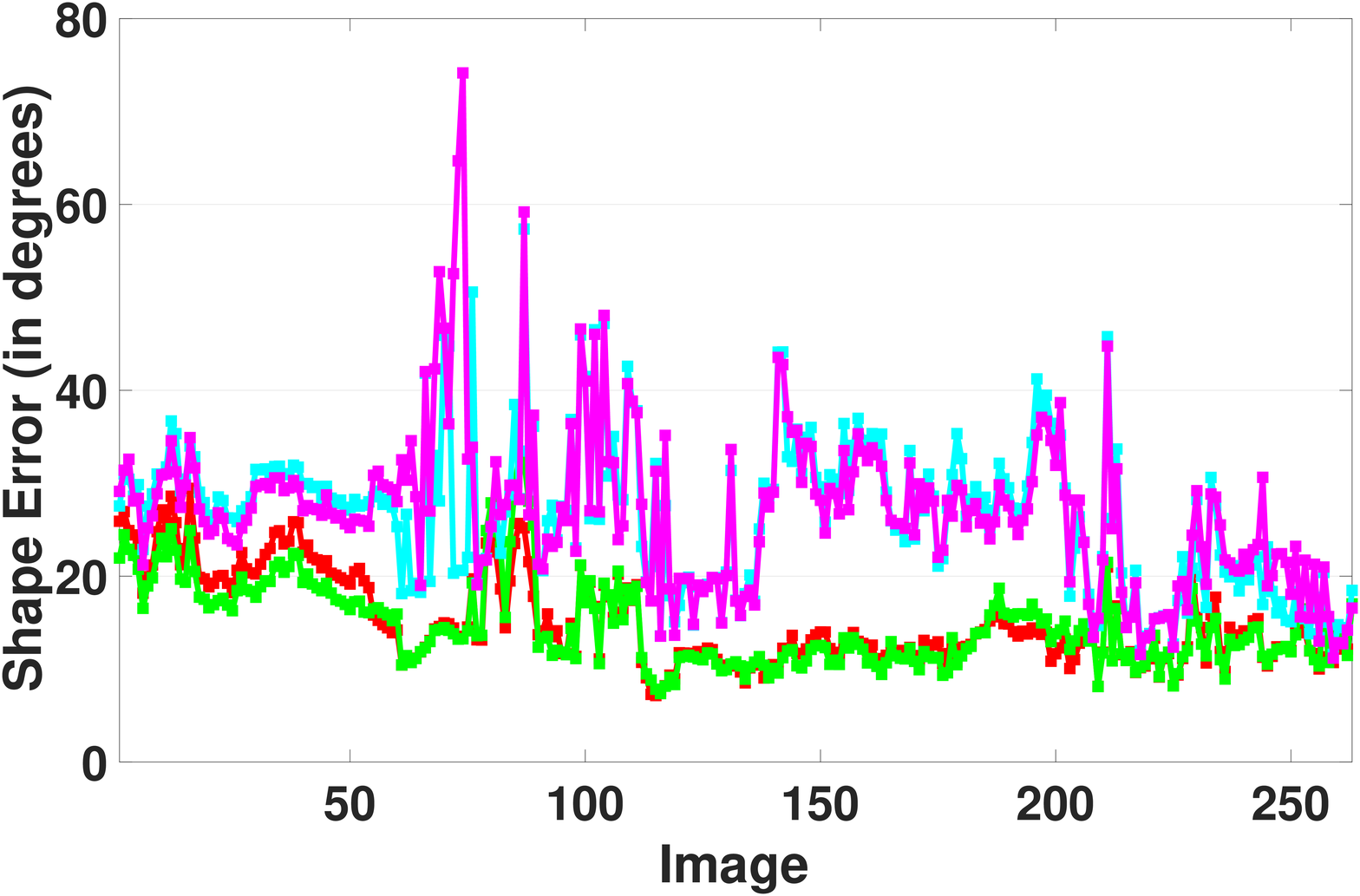}
\includegraphics[width=0.47\textwidth]{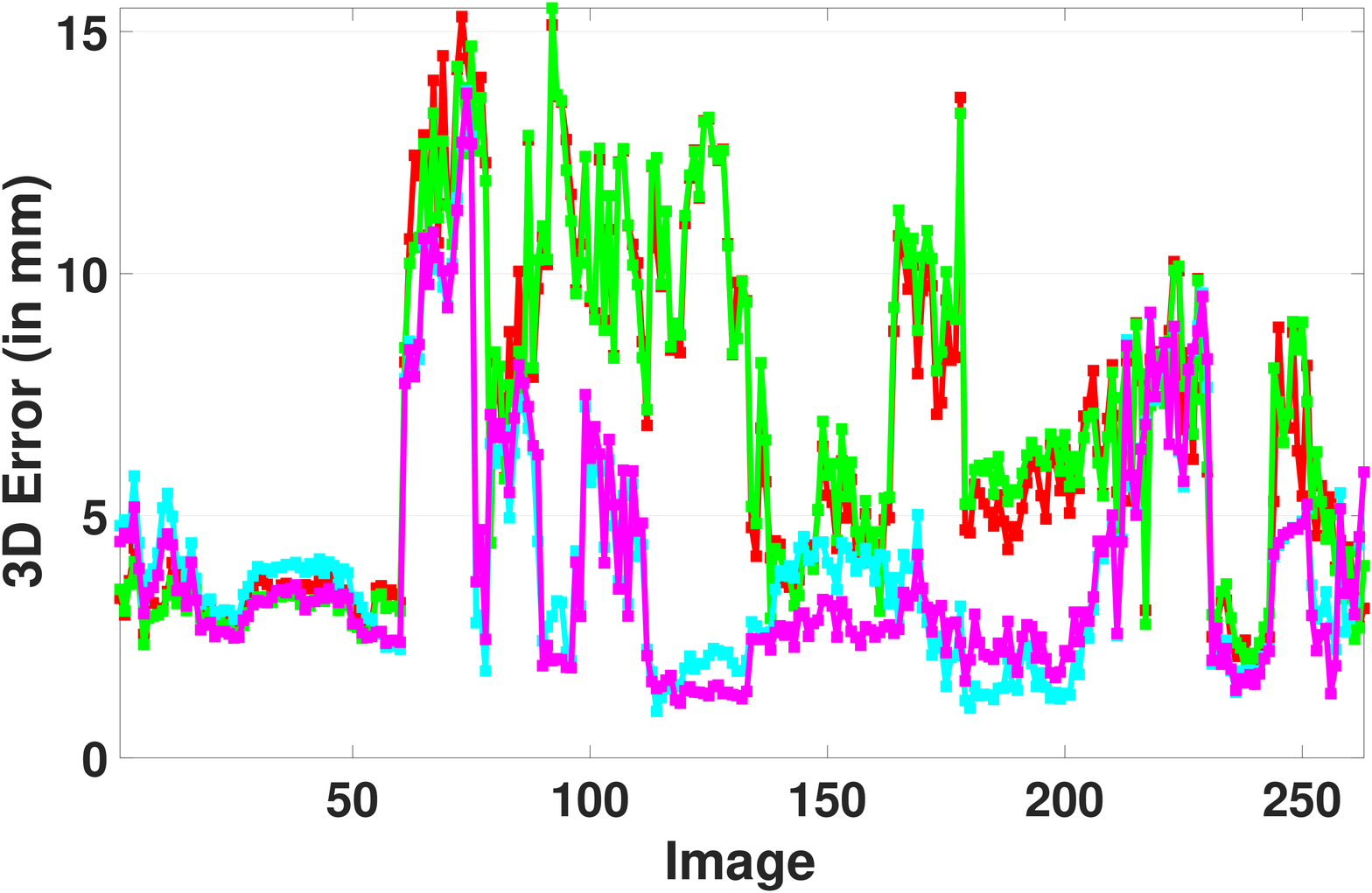}
\includegraphics[width=0.47\textwidth]{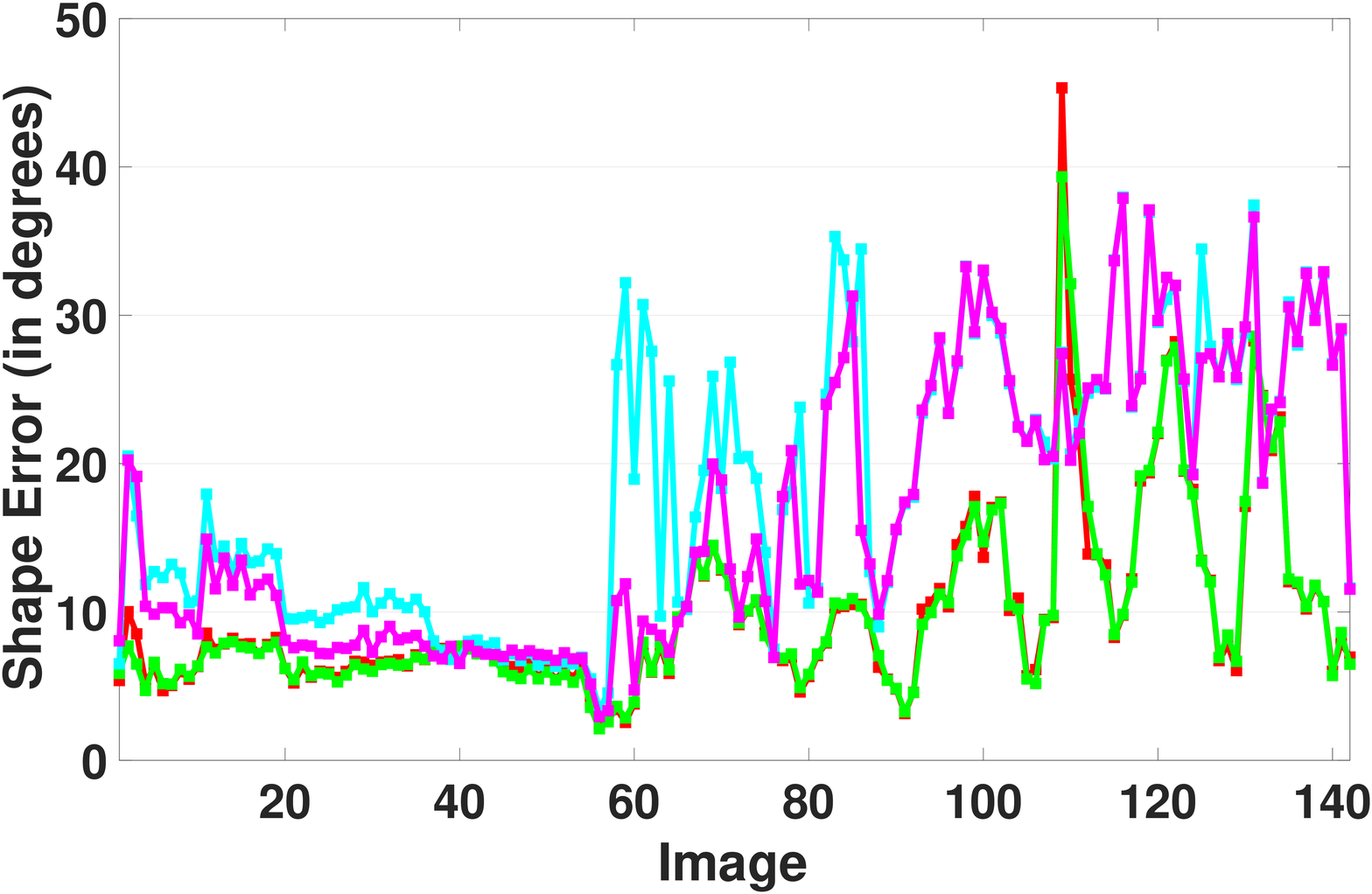}
\includegraphics[width=0.47\textwidth]{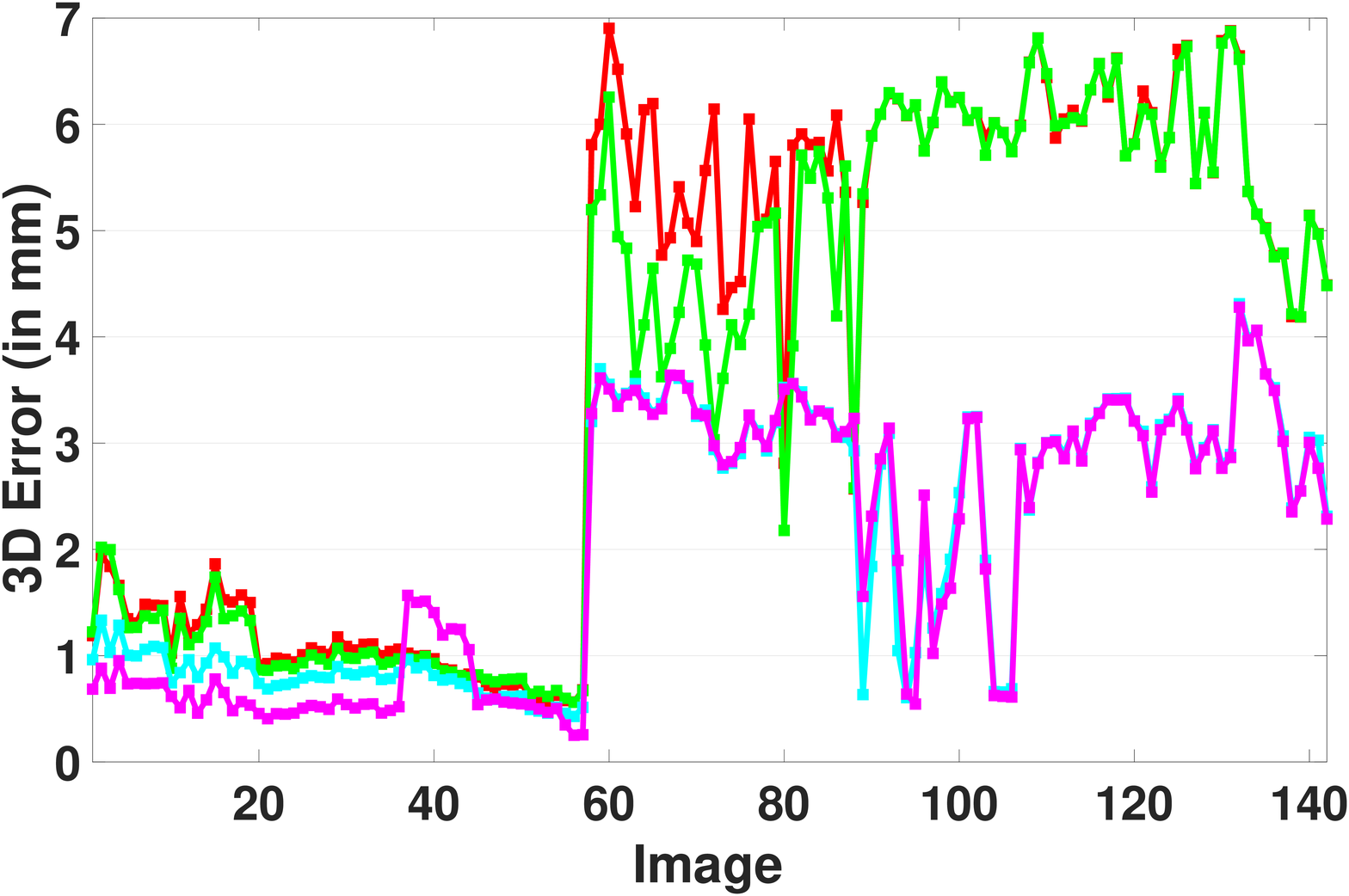}
\includegraphics[width=0.40\textwidth]{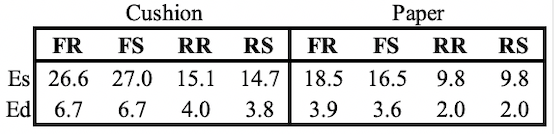}
\caption{Shape and depth errors for the cushion (top) and the paper (bottom) in the NRSfM use-case. The table summarises the mean shape errors (Es) and mean depth errors (Ed) for \textbf{FR-NRSFM} (\textbf{FR}), \textbf{FS-NRSFM} (\textbf{FS}), \textbf{RR-NRSFM} (\textbf{RR}) and \textbf{RS-NRSFM} (\textbf{RS}) over the entire data.}
\label{fig:a}
\end{figure}

\noindent \textbf{{\em With} Correspondence Errors: \textit{Rug}~\cite{Parashar2017}, \textit{ASIFT-Tshirt} and \textit{ASIFT-Hulk}.} 
The \textit{Rug} dataset has $159$ short-baseline images of a  $1$ m wide deforming carpet with $3900$ correspondences obtained from optical flow~\cite{Garg2013}, containing many correspondence errors of small amplitude caused by drift. We defined the ground truth for the correspondences manually and found that at least one-third of the images that appear towards the end of this video are affected by the drift. Many of these affected images contain up to $70\%$ of correspondences affected by an average drift of $20$ pixels. \textit{ASIFT-Tshirt} and \textit{ASIFT-Hulk} are formed of the same images as the original datasets~\cite{Ajad2014} but with correspondences obtained with  ASIFT~\cite{yu2011}.
These correspondences may contain large errors because of the weakly discriminative local texture, see figure~\ref{fig:4} (b). Also, there is approximately $95\%$ missing data in these correspondences.

\textit{Rug} is impacted by optical drift. Figure~\ref{fig:5} (a) shows that \textbf{Go11} has large errors. \textbf{Da12} and \textbf{Le16} perform slightly better than \textbf{Go11}. We could not manage to run \textbf{Va09} and \textbf{Ch14}  because of the large number of images. \textbf{Ch17} and \textbf{Pa17} show a similar performance. \textbf{Pa18} performs worse than \textbf{Pa17}. It is very close to \textbf{Go11}. For \textit{ASIFT-Tshirt} and \textit{ASIFT-Hulk}, the performance of all state-of-the-art methods degrade significantly. \textbf{Ch17}, which showed the best performance with manual correspondences, cannot handle such large amount of missing data. It does not draw enough constraints and thus fails to perform reconstruction.
\textbf{Va09} also cannot cope with such a large amount of missing data and fails.
\textbf{Pa17}, \textbf{Pa18} and \textbf{Ch17} do not perform well. \textbf{FS-NRSfM} and \textbf{FR-NRSfM} also do not show a good performance, being only slightly better than \textbf{Pa17} and \textbf{Ch17}.

\begin{figure*}
\centering
\includegraphics[width=\textwidth]{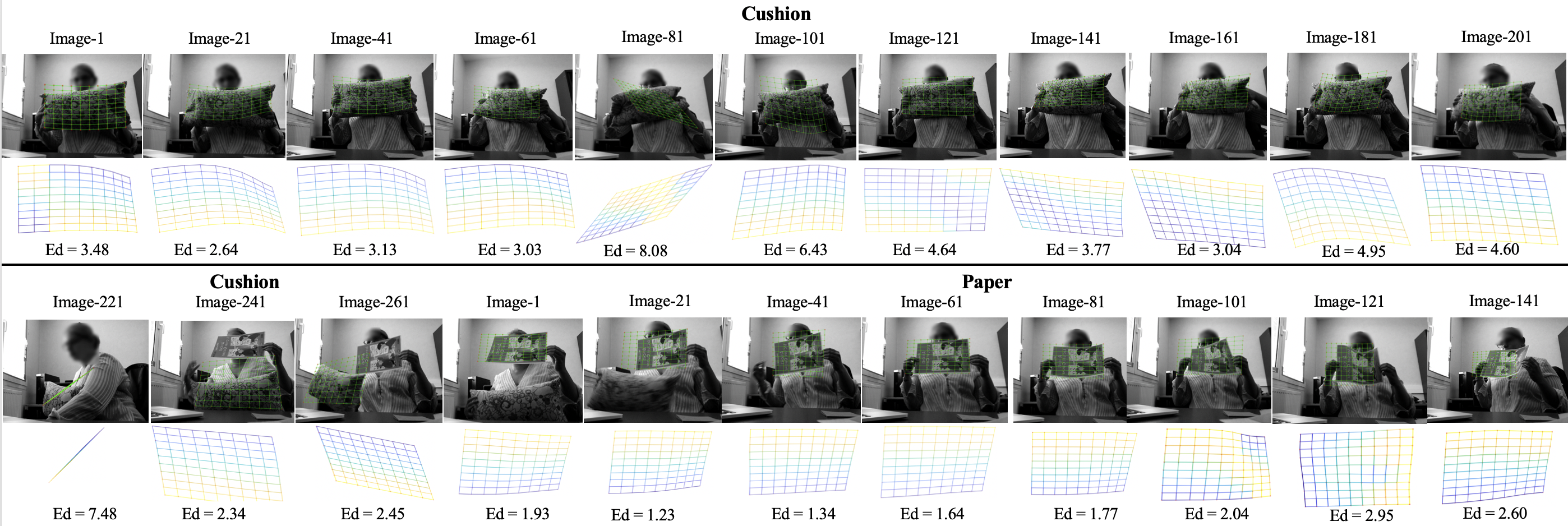}
\caption{Reconstruction of Cushion and Paper from the uniformly sampled NRSfM use-case sequence using \textbf{RR-NRSfM}. A grid computed from reconstructed points is shown to display the geometry of the 3D object. The color of the grid points represent the relative depth of the grid points, blue being closest to the camera and yellow being the farthest. Ed represents the mean depth error measured in $mm$.}
\label{fig:c}
\end{figure*}

In \textit{Rug}, \textbf{MVS} and \textbf{MVR} perform better than their baselines, \textbf{FS-NRSfM} and  \textbf{FR-NRSfM}. Both methods perform slightly better when combined with \textbf{MAD} and \textbf{OR}. 
In  \textit{ASIFT-Tshirt} and \textit{ASIFT-Hulk}, the performances of \textbf{FS-NRSfM} and \textbf{FR-NRSfM} improve when combined with \textbf{MAD} and \textbf{OR}, however it still remains either unsatisfactory or only marginally satisfactory. \textbf{MVS} and \textbf{MVR} shows significantly better results, which are further improved with the use of \textbf{MAD} and \textbf{OR}.
While the use of \textbf{OR} alone on these three  datasets  sacrifices almost $15\%$ 3D points on the entire dataset, the use of \textbf{OR} along with \textbf{MAD}  shows similar or slightly better results and removes fewer than $0.1\%$ 3D points.
 Figure~\ref{fig:5} (b) shows the ablation study and~\ref{fig:5} (c)  shows the ROC analysis of our robust pipeline with \textbf{FR-NRSfM} as the base method. The results are very similar with \textbf{FS-NRSfM} and therefore, we did not report them. 
 Figure~\ref{fig:4} shows the distribution of mismatches on one of the images in  \textit{Rug} and  \textit{ASIFT-Tshirt}  and the distribution of the detected outliers with and without \textbf{MAD}. 
 In  \textit{Rug}, most of false correspondences lie in the low error range of $10$-$20$ pixels.  
 Figure~\ref{fig:4} (a) shows that without \textbf{MAD}, \textbf{OR} does remove a larger number of false 3D points  but at the cost of losing a large number of 3D points which is not preferable. However in \textit{ASIFT-Tshirt}, correspondence errors are much larger.
 Figure~\ref{fig:4} (b) shows that  \textbf{MAD} correctly classifies a larger number of false 3D points  and  drastically reduces the falsely classified 3D points.
 Overall, we observe that \textbf{RR-NRSfM} and \textbf{RS-NRSfM} significantly improve the results.

To summarize, 
our robust methods perform significantly better than other methods in  datasets with errors, whereas they are very close to the best performing  method on datasets without errors. We found that each step in our robust NRSfM pipeline is important. \textbf{MVR}/\textbf{MVS} allows us to reconstruct from the best ordered set of images while \textbf{OR} combined with \textbf{MAD} achieves an intelligent detection of outliers which is key  to achieve robustness.  Both \textbf{RS-NRSfM} and \textbf{RR-NRSfM} show a similar performance. \textbf{RS-NRSfM} is slightly computationally cheaper than \textbf{RR-NRSfM} as it does not have to compute resultant to create the univariate polynomial. On the other hand, \textbf{RR-NRSfM} is slightly more accurate as it prunes insignificant coefficients of the equations in order to compute the resultants which makes the equations well conditioned. 

\begin{figure*}
\centering
\includegraphics[width=\textwidth]{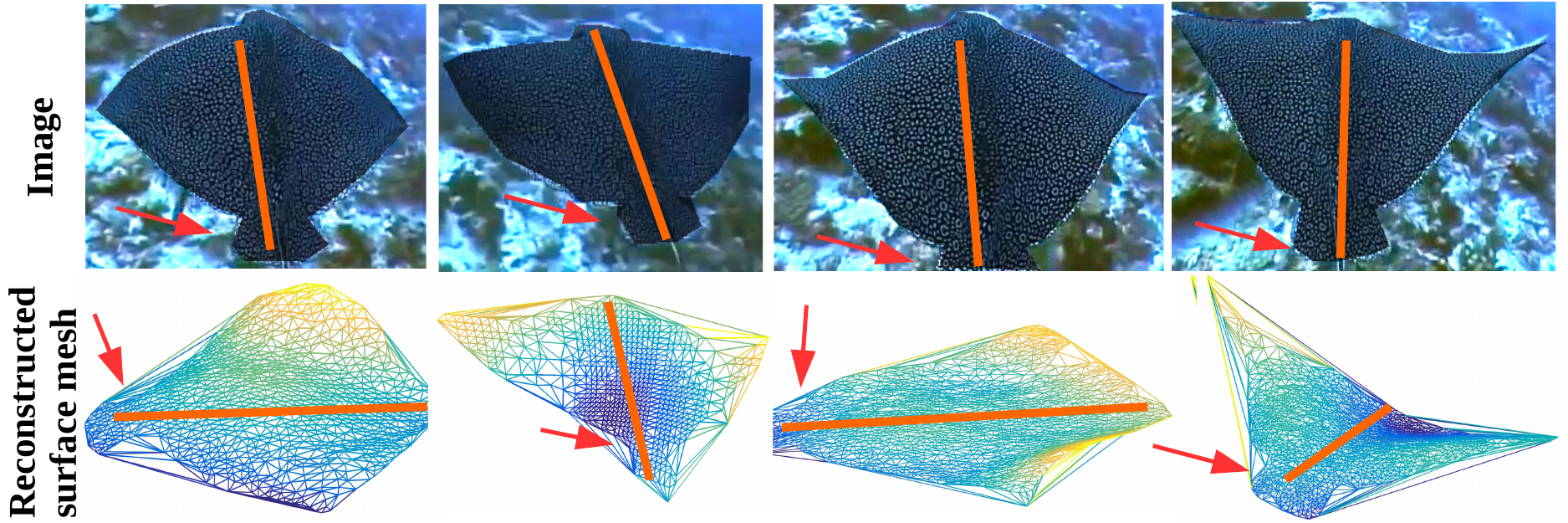}
\caption{The eagle-ray footage with \textbf{RR-NRSfM} reconstruction. The red arrow indicates the tail and the orange line indicates the spine of the eagle-ray.}
\label{fig:b}
\end{figure*}

\noindent \textbf{NRSfM Use-case.} 
We introduce a video of two isometrically deforming objects, a $40 cm$ long cushion  and a $20 cm$ paper sheet, depicting a realistic interaction of objects in the real world as shown in figure~\ref{fig:a}. The ground truth is obtained using Kinect. The user picks the two objects, one by one, and deforms them in front of the camera.
The user provides as input an object bounding box in one frame from which we start tracking it using ASIFT.
While the user is picking up and dropping an object, they may suffer from occlusions and motion blur. 
There are frames in which none or both of the objects are visible. At some point, the user keeps the object still which introduces degeneracies. These conditions are avoided in the usual datasets used to evaluate NRSfM. However, they are extremely important to deal with as capturing a deforming object and avoiding these conditions is unrealistic and impractical. 
\textbf{Go11}, \textbf{Da12}, \textbf{Le16}, \textbf{Ch17}, \textbf{Ch14}, \textbf{Va09}, \textbf{Pa18} and \textbf{Pa17} failed. Figure~\ref{fig:a} shows that \textbf{FR-NRSfM} and \textbf{FS-NRSfM} did not perform well for both objects. \textbf{RR-NRSfM} and \textbf{RS-NRSfM} performed better than \textbf{FR-NRSfM} and \textbf{FS-NRSfM}, with \textbf{RR-NRSfM} being slightly better than \textbf{RS-NRSfM}.  The spikes in the graphs of figure~\ref{fig:a} for both objects arise due to the object remaining still, occlusions and motion blur as the user picks up or puts down the objects. Our robust methods \textbf{RR-NRSfM} and \textbf{RS-NRSfM} successfully reconstruct all frames without failing even if the viewing conditions are not favorable and re-establish  accuracy once the object is viewed in better conditions. Figure~\ref{fig:c} shows the reconstruction of the Cushion and Paper using \textbf{RR-NRSfM} on the use-case sequence. The images are uniformly sampled images from the entire sequence.

\noindent \textbf{Eagle-ray Footage.}
We downloaded a low resolution $360\times480$ video from Youtube. We computed the camera intrinsics using  Structure-from-Motion~\cite{PhotoScan}   with points in the background and correspondences on the deformable eagle-ray using optical flow~\cite{Sundaram2010}. \textbf{Go11}, \textbf{Da12}, \textbf{Le16}, \textbf{Ch17}, \textbf{Ch14}, \textbf{Va09}, \textbf{Pa18} and \textbf{Pa17} failed. \textbf{FR-NRSfM} and \textbf{FS-NRSfM} did not  perform well  as most correspondences are short-ranged. \textbf{RR-NRSfM} and \textbf{RS-NRSfM} perform similarly well, some of the reconstructions using \textbf{RR-NRSfM} are shown in figure~\ref{fig:b}.
 
\noindent \textbf{Discussion.} The first step in our robust NRSfM pipeline is \textbf{MAD} which identifies the potential mismatches between image correspondences.~\cite{Tran2012} is one of the existing methods that performs an outlier-rejection on image correspondences. We replaced \textbf{MAD} with this method in our robust NRSfM pipeline and reconstructed the ASIFT-Tshirt and ASIFT-Hulk datasets. In doing so, the mean depth error on these datasets dropped slightly from 8.3 $mm$ and 7.5 $mm$ to 6.7 $mm$ and 6.2 $mm$ respectively. However,~\cite{Tran2012} takes around 500 ms  per image to identify mismatches whereas \textbf{MAD} does so in only 150-200 ms. Therefore we did not use~\cite{Tran2012} in further experiments.

\section{Conclusions}
We have presented the first entirely statistically robust NRSfM pipeline for near-isometric objects. It computes the local optical flow at each correspondence in the images, reconstructs the 3D point clouds up-to-scale and performs isometry consistent rescaling. It identifies inliers and outliers and thus deals with correspondence errors effectively. We have also introduced two new fast correspondence-wise solutions to NRSfM that estimate the normals locally and guarantee a real solution. Experimental results on synthetic and real datasets showed that our robust methods outperform the existing ones in both accuracy and robustness. 
They are thus the first truly robust NRSfM methods, in the sense of their ability to reject false correspondences.
With them, we were able to drop the manual error-free correspondences of a public, wide-baseline dataset, and to use instead automatically established correspondences using an off-the-shelf ASIFT implementation.
This is a significant step forward in making NRSfM able to deal with realistic datasets in full autonomy.
In future work, we plan to explore the application of our method incrementally in realtime.






\bibliographystyle{ieee}
\bibliography{egbib}
%
%
%

%

\begin{IEEEbiography}
[{\includegraphics[width=1in,height=1.25in,clip,keepaspectratio]{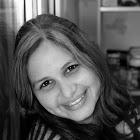}}]{Shaifali Parashar}
  received her PhD degree in Computer Vision from Universit\'e Clermont Auvergne. She is currently a Post-doc researcher in CVLAB, EPFL. Her research interest are non-rigid 3D reconstruction, deformable SLAM and 3D shape registration.
\end{IEEEbiography}

\begin{IEEEbiography}[{\includegraphics[width=1in,height=1.25in,clip,keepaspectratio]{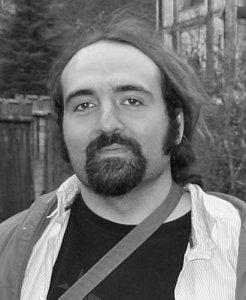}}]{Daniel Pizarro}
has been an Associate Professor at the Universidad de Alcal\'a (Spain) since 2012. He is a member of the GEINTRA group and an invited member of EnCoV group. His research interests include image registration, deformable reconstruction and their applications to minimally invasive surgeries.
\end{IEEEbiography}

\begin{IEEEbiography}[{\includegraphics[width=1in,height=1.25in,clip,keepaspectratio]{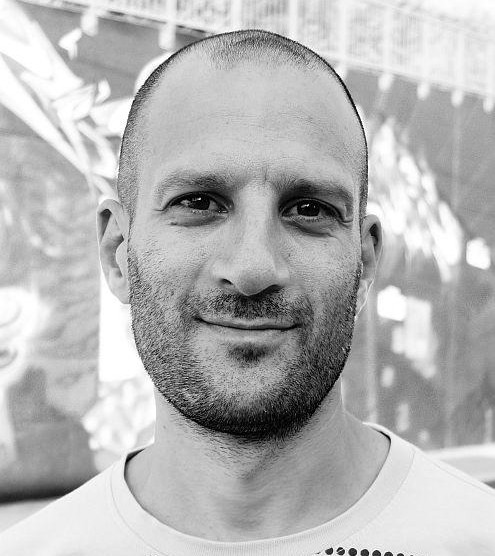}}]{Adrien Bartoli}
has been a Professor of Computer Science at Universit\'e Clermont Auvergne since fall 2009.
He leads the EnCoV research group at Institut Pascal (CNRS, UCA, CHU de Clermont-Ferrand).
His research interests include image registration and Shape-from-X for rigid and non-rigid environments, with applications to computer-aided endoscopy.
\end{IEEEbiography}



\clearpage
\appendices
\begin{appendices}
 \section{Coefficients of the Reconstruction Equations}
 \label{appendixA}
 The coefficients of the two cubic reconstruction equations~\eqref{eq:eq1} and~\eqref{eq:eq2} are:
 
 \begin{align}
\nonumber
 a_{30}  &\!=\! 2 e_{1} j_1 {j_3}^2e_{10} \!-\! 2 e_{1} j_1^2 {j_3}e_9\!+\!2e_2j_1j_3v_1e_{16}, \\ \nonumber 
a_{21} & \!=\! 2e_{1}j_3(j_1j_4\!+\!e_{15})e_{10}\!-\!2e_1j_1(e_{15}\!+\!j_2j_3)e_9 \\ \nonumber &\!+\!2e_{16}e_2(v_1e_{15}\!-\!j_1j_3u_1),\\ \nonumber 
a_{12}  &\!=\! 2e_{1}j_4(e_{15}\!+\!j_2j_3)e_{10}\!-\!2e_1j_2(j_1j_4\!+\!e_{16})e_9 \\ \nonumber &\!+\!2e_2e_{16}(j_2j_4v_1\!-\!u_1e_{15}), 
\\ \nonumber 
a_{03}  &\!=\! 2 e_{1} j_2 {j_4}^2e_{10} \!-\! 2 e_{1} j_2^2 {j_4}e_9\!-\!2e_2j_2j_4u_1e_{16}, 
 \\ \nonumber 
a_{20}  &\!=\! 4j_1j_3e_3e_9\!-\!4j_1j_3e_4e_{10}\!+\!j_1^2e_1e_5\!-\!j_3^2e_1e_6\!-\!e_2e_{15}e_{16}, \\ \nonumber 
a_{11}  &\!=\! 4(e_3e_{15}e_9\!-\!e_4e_{16}e_{10})\!+\!2e_1(j_1j_2e_5\!-\!j_3j_4e_6)\\ \nonumber &\! +\!2e_2(j_1j_3\!-\!j_2j_4)e_{16}, \\ \nonumber 
a_{02}  &\!=\! 4j_2j_4e_3e_9\!-\!4j_2j_4e_4e_{10}\!+\!e_1j_2^2e_5\!+\!e_1j_4^2e_6\!+\!e_2e_{15}e_{16}, \\ \nonumber 
a_{10}  &\!=\!  2j_1e_{14}e_{10}\!-\!2j_3e_{13}e_9\!-\!2j_1e_3e_5\!+\!2j_3e_4e_6 , \\ \nonumber 
a_{01}  &\!=\! 2j_2e_{14}e_{10}\!-\!2j_4e_{13}e_9\!-\!2j_2e_3e_5\!+\!2j_4e_4e_6 , \\ \nonumber 
a_{00}  &\!=\! e_{13}e_5\!-\!e_{14}e_6 , \text{ }
b_{30}  \!=\! e_1j_3^3e_{10}\!-\!j_1j_3^2e_1e_9\!+\!j_3^2e_2v_1e_{16}, \\ \nonumber 
b_{21}  &\!=\! 3j_3^2j_4e_1e_{10}\!-\!j_3e_1(j_1j_4\!+\!e_{15})e_9\!+\!j_3e_2(2j_4v_1\!-\!j_3u_1)e_{16} , \\ \nonumber 
b_{12}  &\!=\! 3j_3j_4^2e_1e_{10}\!-\!j_4e_1(e_{15}\!+\!j_2j_3)e_9\!+\!j_4e_2(j_4v_1\!-\!2j_3u_1)e_{16}, \\ \nonumber 
b_{03}  &\!=\! e_1j_4^3e_{10}\!-\!j_2j_4^2e_1e_9\!+\!j_4^2e_2u_1e_{16}, \\ \nonumber 
b_{20}  &\!=\! j _1j_3e_1e_5\!+\!2j_3^2e_3e_9\!-\!2j_3^2e_4e_{10}\!+\!j_3^2e_1e_8\!-\!j_3j_4e_2e_{16}, \\ \nonumber 
b_{11}  &\!=\! e_{15}e_1e_5\!+\!2j_3j_4(2e_3e_9\!-\!2j_4v_1e_{10}\!+\!e_1e_8)\!+\!e_{16}e_2(j_3^2\!-\!j_4^2), \\ \nonumber 
b_{02}  &\!=\! j _2j_4e_1e_5\!+\!2j_4^2e_3e_9\!-\!2j_4^2e_4e_{10}\!+\!j_4^2e_1e_8\!+\!j_3j_4e_2e_{16}, \\ \nonumber 
b_{10}  &\!=\! j_3e_{14}e_{10} \!-\!(j_3e_7\!-\!j_4e_{16})e_9\!-\!(j_1e_4\!+\!j_3e_3)e_5\!-\!2j_3e_4e_8
, \\ \nonumber 
b_{01}  &\!=\! j_4e_{14}e_{10} \!-\!(j_3e_{16}\!-\!j_4e_7)e_9\!-\!(j_4e_3\!+\!j_2e_4)e_5\!-\!2j_4e_4e_8
, \\ \nonumber 
b_{00}  &\!=\!e_{14}e_8\!+\!e_7e_5, \text{ }
e_1 \!=\! 1 \!+\! u_1^2 \!+\! v_1^2, \text{ } e_2 \!=\! 1 \!+\! u_2^2 \!+\! v_2^2,    \\ \nonumber 
e_3 &\!=\! j_1u_1\!+\!j_2v_1, \text{ }e_4 \!=\! j_3u_1\!+\!j_4v_1, \text{ } e_5\! = \!1\!-\!2t_2v_2\!+\!e_2t_2^2,  \\ \nonumber  e_6 &\!= \!1\!-\!2t_1u_2\!+\!e_2t_1^2, 
e_7 \!=\!j_1j_3\!+\!j_2j_4,  e_8 \!=\!t_2u_2\!+\!t_1v_2\!-\!e_2t_1t_2,\\ \nonumber 
e_{9}  &\!= v_2\!\!-\!e_2t_2,  \text{ }
e_{10}  \!= u_2\!\!-\!e_2t_1,  \text{ }
e_{11}  \!= j_2u_1\!\!+\!j_4v_1, \\ \nonumber 
e_{12} &\!=\! j_2u_1\!-\!j_3u_1, \text{ }
e_{13} \!=\! j_1^2\!+\!j_2^2, \text{ } e_{14}\! =\! j_3^2\!+\!j_4^2, \text{ } e_{15} \!=\! j_1j_4 \!+\!j_2j_3,\\ \nonumber
e_{16} &\!=\! j_1j_4 \!-\!j_2j_3, \text{ }t_1 \!=\! \!-\!(j_3h_3\!+\!j_4h_4), \text{ } 
t_2 \!=\! \!-\!(j_1h_3\!+\!j_2h_4) .
\end{align}

\end{appendices}

\end{document}